\documentclass{article}
\usepackage[a4paper, total={6in, 8in}]{geometry}
\usepackage{graphicx} % Required for inserting images

\usepackage[round]{natbib}
\usepackage{amsmath,amsfonts}
\usepackage[affil-it]{authblk}
\usepackage{multirow}
\usepackage{comment}
\usepackage{wrapfig}
\usepackage{xcolor}
\usepackage{tcolorbox}
\definecolor{darkpink}{rgb}{0.91, 0.33, 0.5}

\usepackage[utf8]{inputenc} % allow utf-8 input
\usepackage[T1]{fontenc}    % use 8-bit T1 fonts
\usepackage{hyperref}       % hyperlinks
\usepackage{url}            % simple URL typesetting
\usepackage{booktabs}       % professional-quality tables
\usepackage{amsfonts}       % blackboard math symbols
\usepackage{nicefrac}       % compact symbols for 1/2, etc.
\usepackage{microtype}      % microtypography
\usepackage{xcolor}         % colors
\usepackage{amsmath}
\usepackage{amssymb}
\usepackage{mathtools}
\usepackage{amsthm}
\usepackage{enumitem}

\usepackage{mdframed}
\usepackage{tikz}
\usetikzlibrary{positioning, arrows, fit, backgrounds}
\usetikzlibrary {shapes.multipart}
\usepackage{epigraph}
\theoremstyle{plain}
\newtheorem{theorem}{Theorem}%[section]

\theoremstyle{definition}
\newtheorem{definition}{Definition}

\theoremstyle{remark}

\DeclareMathAlphabet{\mathcal}{OMS}{cmsy}{m}{n}
\SetMathAlphabet{\mathcal}{bold}{OMS}{cmsy}{b}{n}

\definecolor{myblue}{RGB}{30, 144, 255}

\title{From superposition to sparse codes: interpretable representations in neural networks}

\author{David Klindt%
    \thanks{Corresponding author: \texttt{klindt@cshl.edu}}}
\affil{
Cold Spring Harbor Laboratory\\
Cold Spring Harbor, NY, USA
}
\author{Charles O'Neill}
\affil{Australian National University, Canberra, AU}
\author{Patrik Reizinger}
\affil{
Max Planck Institute for Intelligent Systems, ELLIS Institute\\
Tübingen, Germany;
}
\author{Harald Maurer}
\affil{
University of Tübingen\\
Tübingen, Germany
}
\author{Nina Miolane}
\affil{
University of California Santa Barbara\\
Santa Barbara, CA, USA
}

\begin{document}

\maketitle

%%%%%%%%%%%%%%%%%%%%%%%%%%%%%%%%%%%%%%%%%%%%%%%%%%%
%%%%%%%%%%%%%%%%%%%%%%%%%%%%%%%%%%%%%%%%%%%%%%%%%%%
\vspace{-20pt}
\section{Abstract}
Understanding how information is represented in neural networks is a fundamental challenge in both neuroscience and artificial intelligence. Despite their nonlinear architectures, recent evidence suggests that neural networks encode features in \textit{superposition}, meaning that input concepts are linearly overlaid within the network’s representations. We present a perspective that explains this phenomenon and provides a foundation for extracting \textit{interpretable representations} from neural activations. Our theoretical framework consists of three steps: (1) \textbf{Identifiability theory} shows that neural networks trained for classification recover latent features up to a linear transformation. (2) \textbf{Sparse coding methods} can extract disentangled features from these representations by leveraging principles from compressed sensing. (3) \textbf{Quantitative interpretability} metrics provide a means to assess the success of these methods, ensuring that extracted features align with human-interpretable concepts. By bridging insights from theoretical neuroscience, representation learning, and interpretability research, we propose an emerging perspective on understanding neural representations in both artificial and biological systems. Our arguments have implications for neural coding theories, AI transparency, and the broader goal of making deep learning models more interpretable.

%%%%%%%%%%%%%%%%%%%%%%%%%%%%%%%%%%%%%%%%%%%%%%%%%%%
%%%%%%%%%%%%%%%%%%%%%%%%%%%%%%%%%%%%%%%%%%%%%%%%%%%

\begin{figure}[htb]
    \centering
    \includegraphics[width=0.99\textwidth]{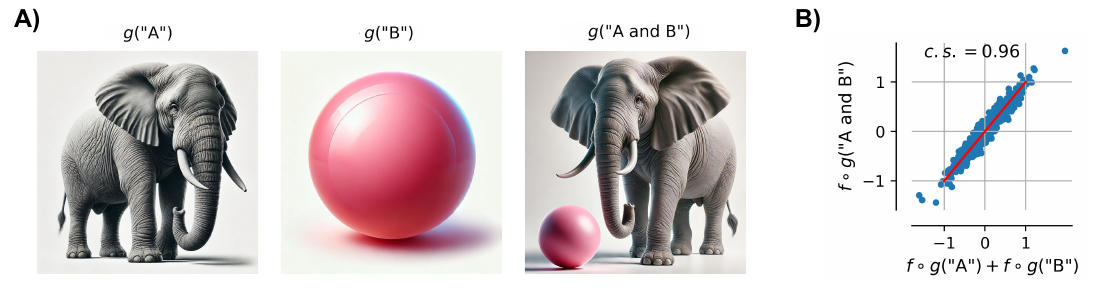}
    \includegraphics[width=0.99\textwidth]{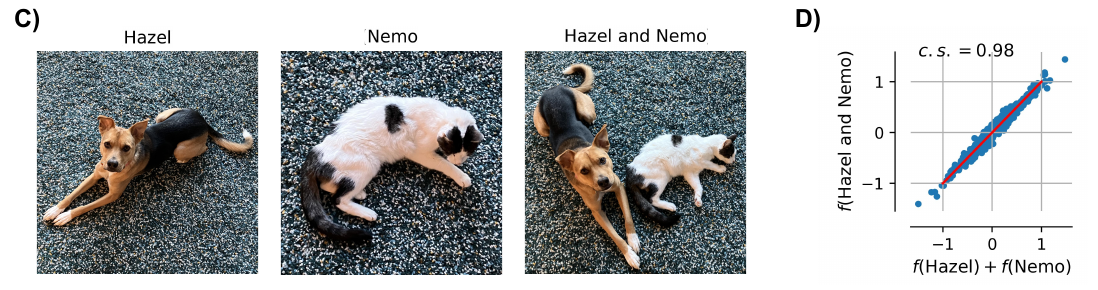}
    \vspace{10pt}
    \caption{
    \textbf{Superposition of neural representations.}
    \textbf{A)} Images generated from GPT-4 prompts: 
    ``A'': \textit{an elephant}, ``B'': \textit{a pink ball}, and ``A and B'': \textit{an elephant and a pink ball};
    \textbf{B)} \textit{Test for additivity:} Adding the neural representations (ViT-B/16 \citep{dosovitskiy_image_2021}) of the first two images ($f \circ g(\text{``A''})$ and $f \circ g(\text{``B''})$) and the neural representation for the combined image ($f \circ g(\text{``A and B''})$) (c.s.: cosine similarity; each dot corresponds to one dimension of the neural representation);
    \textbf{C)} Real (not AI generated) images of a dog (``Hazel''), a cat (``Nemo'') and both together.
    \textbf{D)} Same analysis as in \textbf{B)}, but with the images in \textbf{C)}, showing that additivity even holds on natural images.
    Further examples ($N=10$, not shown) support that this finding is statistically significant (over a calibrated baseline, i.e., the c.s.~of the representations of just ``A'' and ``B'') in neural representation space ($f \circ g$, $p < 10^{-5}$) but not in pixel space ($g$, $p \approx 0.88$).
    }
    \label{fig:cnn_superposition}
    \vspace{10pt}
\end{figure}

%%%%%%%%%%%%%%%%%%%%%%%%%%%%%%%%%%%%%%%%%%%%%%%%%%%
%%%%%%%%%%%%%%%%%%%%%%%%%%%%%%%%%%%%%%%%%%%%%%%%%%%
\section{Main}
Understanding how information is represented in neural networks is a fundamental challenge in both neuroscience and artificial intelligence (AI) \citep{helmholtz1878,mcculloch1943logical,kleene1951representationof,rosenblatt1958perceptron,smolensky_tensor_1990}. As AI systems—particularly large language models (LLMs)—become integral to critical domains like healthcare, finance, and security, the need for transparency and interpretability is more urgent than ever \citep{linardatos2020explainable,fan2021interpretability,gilpin2018explaining,leavitt_towards_2020}. 
Neural networks, both biological and artificial, encode information in \textit{distributed representations}, where patterns of activity across many units encode complex concepts \citep{rumelhart_general_1986}. In human cognition, symbolic knowledge emerges from this distributed processing \citep{fodor_language_1975, chomsky2014minimalist, tenenbaum2011grow}. 
% For instance, the discovery of a new species begins with raw sensory data (images, sounds, samples) before being distilled into a symbolic representation (e.g., a Wikipedia entry). 
Similarly, deep learning models generate high-dimensional representations of input data, whether in vision, language, or multimodal systems. But even prior, from classical methods such as Latent Semantic Analysis (LSA) to more recent word2vec \citep{mikolov_efficient_2013} and GloVe \citep{pennington_glove_2014}, all produce distributed word embeddings by capturing semantic relationships through vector operations. 
This raises a fundamental question:

\begin{center}
    \textit{How can we relate distributed neural representations back to human-interpretable concepts?}
\end{center}

%%%%%%%%%%%%%%%%%%%%%%%%%%%%%%%%%%%%%%%%%%%%%%%%%%%
%%%%%%%%%%%%%%%%%%%%%%%%%%%%%%%%%%%%%%%%%%%%%%%%%%%

\begin{figure}[htb]
    \centering
    \includegraphics[width=0.83\textwidth]{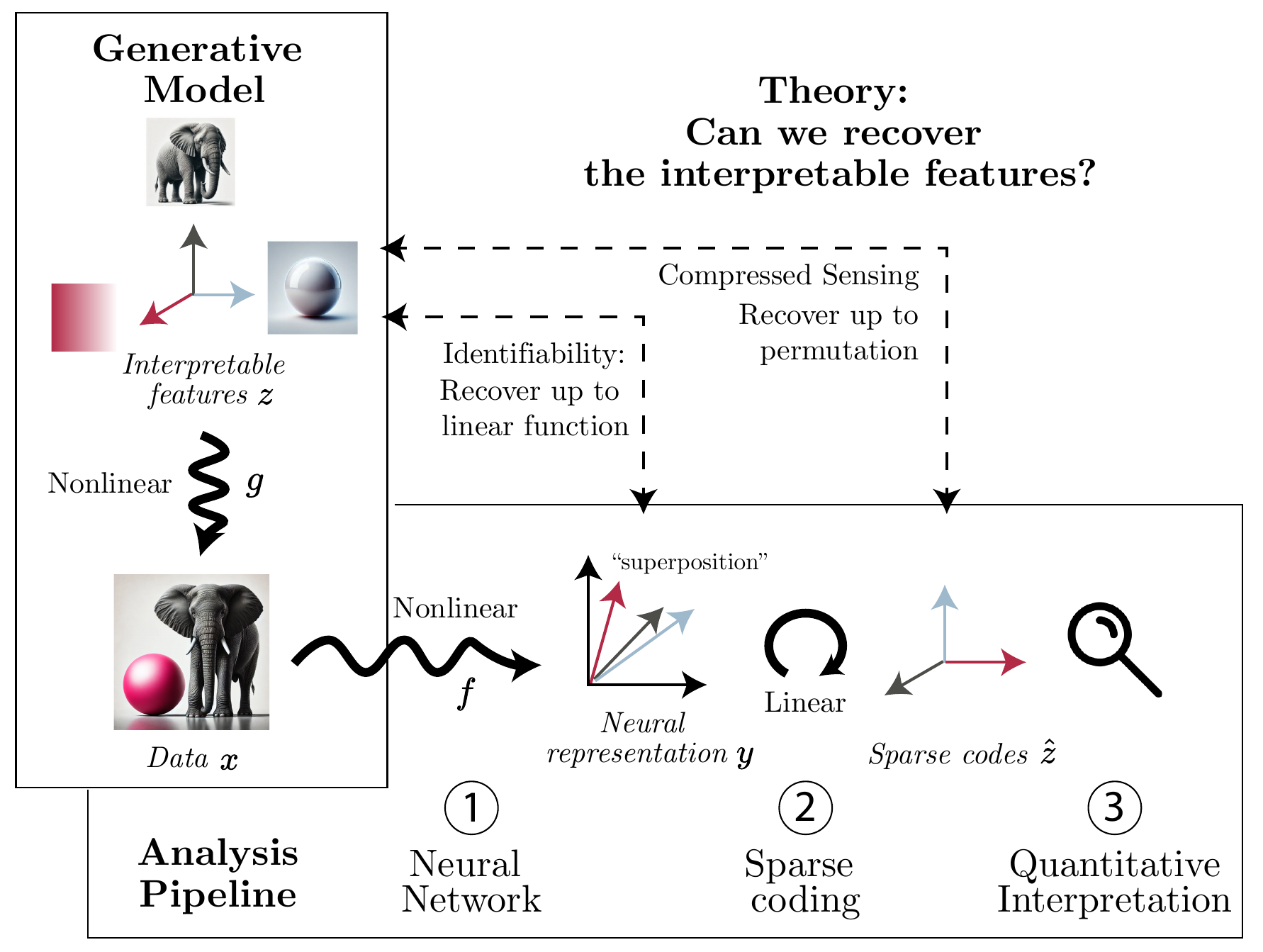}
    \caption{
    \textbf{Theory and analysis pipeline.}
    Data ($x$) arise from interpretable features ($z \in \mathbb{R}^N$) through a nonlinear function ($g$) and neural representations ($y \in \mathbb{R}^N$) arise from data through another nonlinear function ($f$). However, because neural representations have lower dimensionality ($M < N$), they overlay interpretable features in \textit{superposition} \cite{elhage_toy_2022}.
    \textbf{1)} Identifiability theory establishes that the overall mapping from interpretable features to neural representations must be linear.
    \textbf{2)} Compressed sensing theory shows that sparse coding can \textit{lift} sparse features out of superposition, recovering the original interpretable features up to permutations.
    \textbf{3)} Since the true interpretable features are unknown, we evaluate the success of sparse coding using a permutation-invariant measure of interpretability as a proxy.
    }
    \label{fig:pipeline}
    \vspace{10pt}
\end{figure}

%%%%%%%%%%%%%%%%%%%%%%%%%%%%%%%%%%%%%%%%%%%%%%%%%%%
%%%%%%%%%%%%%%%%%%%%%%%%%%%%%%%%%%%%%%%%%%%%%%%%%%%
\vspace{10pt}
\paragraph{The superposition principle.}
A key hypothesis is that meaningful concepts are encoded in \textit{superposition}—where multiple features are \textit{linearly combined} in a shared representational space \citep{van_gelder_what_1990, arora_linear_2018, elhage_toy_2022}. This implies that the representation of a combined concept is well approximated by the sum of the representations of its parts. For instance, as shown in Fig.~\ref{fig:cnn_superposition}, the sum of neural representations for an \textit{elephant} and a \textit{pink ball} approximates the representation of an image containing both. 
At first glance, this is surprising: a highly nonlinear neural network (generative model) transforms inputs into images and then another highly nonlinear neural network (image embedding) transforms inputs into neural representations, yet \textit{together, these transformations result in an additive representation}.
% We argue that this phenomenon fundamentally shapes both neuroscience and AI.
Historically, research on neural representations has distinguished between \textit{local coding} (a single neuron encodes a concept) and \textit{distributed coding} (a concept is represented by patterns across multiple neurons) \citep{rumelhart_general_1986, thorpe_local_1989, quian_invariant_2005}. 
% Superposition provides a bridge between these views: rather than discrete neuron-concept mappings, representations exist in \textit{high-dimensional spaces with structured, additive components}.
\citet{smolensky_tensor_1990} formally defined the \textit{superposition principle}:

% % \begin{mdframed}
% \begin{definition}[Smolensky superposition]\label{def:sup_smol}
%     A `connectionist' [distributed] representation function $\Psi$ employs the superpositional representation of conjunction ($\wedge$) if and only if:
%     \begin{equation}\label{eq:smolensky}
%         \Psi \left(\underset{i}{\bigwedge} \, p_i \right) = \sum_i \Psi(p_i).
%     \end{equation}
% \end{definition}
% % \end{mdframed}
% \noindent
% where $\bigwedge_i p_i$ denotes the logical conjunction of propositions $p_i$ (e.g., \(\text{``A''} \wedge \text{``B''} = \text{``A and B''}\) in Fig.~\ref{fig:cnn_superposition}).

\begin{mdframed}[nobreak=true]
\begin{definition}[Smolensky superposition]\label{def:sup_smol}
    A `connectionist' [distributed] representation function $\Psi$ employs the superpositional representation of conjunction ($\wedge$) if and only if:
    \begin{equation}\label{eq:smolensky}
        \Psi \left(\underset{i}{\bigwedge} \, p_i \right) = \sum_i \Psi(p_i).
    \end{equation}
    where $\bigwedge_i p_i$ denotes the conjunction of propositions $p_i$ (e.g., \(\text{``A''} \wedge \text{``B''} = \text{``A and B''}\) in Fig.~\ref{fig:cnn_superposition}).
\end{definition}
\end{mdframed}
\vspace{10pt}

\noindent
This implies that the representation of a composite concept is the sum of its individual representations (e.g., $\Psi := f \circ g$ in Fig.~\ref{fig:cnn_superposition}). This \textit{additivity} property (\ref{eq:additivity}) suggests that neural networks encode features \textit{linearly in high-dimensional spaces}.
The problem is that such superposed representations are difficult to interpret directly.
\textit{Sparse coding} offers a potential solution: by transforming neural activations into sparse codes, we may uncover representations that better align with human-interpretable concepts \citep{olshausen_emergence_1996}.
In this perspective, we outline a three-step approach for extracting interpretable codes from distributed neural representations:
\vspace{10pt}
\begin{tcolorbox}[colback=blue!5!white, colframe=blue!75!black, label=sec:pipeline, title=\textbf{Theory and analysis pipeline (see also Fig.\ref{fig:pipeline}).}]
\begin{enumerate}
    \item \textbf{Identifiability theory:} Neural networks trained on structured data (provably) \textit{learn to represent latent variables up to a linear transformation}.
    \item \textbf{Sparse coding:} Compressed sensing theory guarantees under what conditions we can extract (interpretable) \textit{sparse codes from superposition}.
    \item \textbf{Quantitative interpretability:} Since true latent variables are unknown, we need interpretability metrics as \textit{proxies to evaluate extracted sparse codes}.
\end{enumerate}
\end{tcolorbox}
\vspace{10pt}

%%%%%%%%%%%%%%%%%%%%%%%%%%%%%%%%%%%%%%%%%%%%%%%%%%%
%%%%%%%%%%%%%%%%%%%%%%%%%%%%%%%%%%%%%%%%%%%%%%%%%%%
\section{Interpretable sparse codes from distributed representations}\label{sec:prior_work}

\epigraph{``It is a capital mistake to theorize before one has data.''}{\textit{Sherlock Holmes}}

\noindent
We now turn to prior research, all of which can be understood from the perspective of our three-step approach: i) assume linear representations, ii) apply sparse coding, and iii) assess interpretability (qualitative and/or quantitative). Table~\ref{tab:prior_results} summarizes key (non-exhaustive) contributions in this area, spanning early matrix factorization techniques to modern deep learning methods.

\begin{table}[ht]
    \small
    \centering
    \begin{tabular}{p{0.26\linewidth} | p{0.11\linewidth} | p{0.15\linewidth} | p{0.39\linewidth}}
        \textbf{Paper} & \textbf{Model} & \textbf{Methods} & \textbf{Finding} \\
        \hline
        \citep{ritter_self-organizing_1989} & SVD & SOM & Qualitative interpretability \\
        \citep{honkela_wordicaemergence_2010} & SVD & ICA & Qualitative interpretability; matches Brown corpus \citep{francis_brown_1979} \\
        \citep{murphy_learning_2012} & SVD & NNSE & Increased interpretability in WIT \\
        \citep{fyshe_compositional_2015} & SVD & compositional NNSE & Similar interpretability as NNSE in WIT, improved phrase prediction \\
        \citep{faruqui_sparse_2015} & GloVe & SC (NN, bin) & Increased interpretability in WIT \\
        \citep{subramanian_spine_2017} & GloVe, word2vec & k-sparse AE & Higher interpretability than \citep{faruqui_sparse_2015} in WIT \\
        \citep{jang_elucidating_2017} & SVD & NNSE & Identification of semantic categories in HyperLex \citep{vulic_hyperlex_2017} \\
        \citep{arora_linear_2018} & GloVe & SC; theory & Increased interpretability in PLT \\
        \citep{yun_transformer_2021} & BERT & SC & Qualitative interpretability \\
        \citep{bricken_towards_2023} & Transformer & SAE & Qualitative interpretability \\
        \citep{templeton_scaling_2024} & LLM & SAE & Qualitative interpretability \\
        \citep{eleutherOpenSource} & LLM & SAE & Automatic quantitative interpretability
    \end{tabular}
    \caption{
    \textbf{Overview of prior results.}
    List of prior research on extracting sparse, interpretable codes from a variety of distributed (neural) representations; While the sparse methods and evaluation frameworks (qualitative vs.~quantitative) have changed over time, there is a growing consensus that his approach yields more interpretable insights into neural codes.
    SVD - singular value decomposition; SOM - self-organizing map \citep{kohonen1982self}; ICA - independent component analysis; NN - non-negative; NNSE - non-negative sparse embedding; SC - sparse coding; bin - binarized; AE - autoencoder; SAE - sparse autoencoder; WIT - Word intrusion task \citep{chang_reading_2009}; PLT - Police lineup task \citep{arora_linear_2018}.
    }
    \label{tab:prior_results}
\end{table}

%%%%%%%%%%%%%%%%%%%%%%%%%%%%%%%%%%%%%%%%%%%%%%%%%%%
%%%%%%%%%%%%%%%%%%%%%%%%%%%%%%%%%%%%%%%%%%%%%%%%%%%
\paragraph{Prior research in AI.}
One of the earliest works, \citep{ritter_self-organizing_1989} used SVD on word co-occurrence statistics, followed by self-organizing maps \citep{kohonen1982self}, to extract qualitatively meaningful concepts from word embeddings.
Other early works aiming to uncover hidden semantic concepts in language representations include WordICA \citep{honkela_emergence_2003,honkela_wordicaemergence_2010} and Latent Dirichlet Allocation (LDA) \citep{blei_latent_2003}. WordICA used Independent Component Analysis (ICA) to enhance the interpretability of dense word representations, while LDA modeled word probabilities based on latent \textit{topics}, providing a statistical method to group words by thematic content \citep{snow_semantic_2006}.
\citet{faruqui_sparse_2015} applied sparse coding to various word embeddings, including GloVe \citep{pennington_glove_2014}, and demonstrated enhanced interpretability in word similarity tasks. Their work showed that human annotators achieved an average accuracy of 57\% on raw embeddings versus 71\% on sparse representations in the Word Intrusion Task (WIT). Furthermore, inter-annotator agreement increased from 70\% to 77\%, indicating more consistent judgments among annotators. Qualitative analysis supported that sparse representations offered clearer, more interpretable groupings.
More recently, \citet{zhang_word_2019} and \citet{yun_transformer_2021} applied sparse coding to contextual word embeddings from models like BERT \citep{devlin_bert_2019}. By processing all transformer layers simultaneously, they could analyze semantic differentiation across layers, extending sparse coding applications to contextualized, sentence-level representations. This approach aligns with recent work on \textit{cross-coders} \citep{lindsey2024crosscoder}. Their qualitative analyses and interactive visualizations further supported the increased interpretability of sparse representations over neuron-aligned embeddings. Testing on the Brown corpus \citep{francis_brown_1979}, they observed more distinct semantic separations in later transformer layers, alongside the detection of patterns they termed `repetitive pattern detectors', possibly akin to \textit{induction heads} discovered later by \citet{olsson_-context_2022}.
Lastly, there has been a surge of recent publications utilizing various forms of Sparse Autoencoders (SAEs) to extract interpretable features from Large Language Models (LLMs) \citep{templeton_scaling_2024,bricken_towards_2023,gao_scaling_2024,lindsey2024crosscoder,lan_sparse_2024,engels_not_2024,oneill_sparse_2024,wu2024openai,rajamanoharan_improving_2024}. Recent work has extended SAE methods to multimodal settings -- applying techniques such as PatchSAE \citep{lim2024sparseautoencodersrevealselective} and universal SAEs \citep{thasarathan2025universalsparseautoencodersinterpretable} to models like CLIP, DinoV2, and even demonstration that SAEs can extract interpretable features that align visual and textual modalities \citep{transformercircuitsScalingMonosemanticity}.
The pace of developments in this area is rapid, and keeping up with each new advancement lies beyond the scope of this work. Below, we discuss the limitations of SAEs in terms of computational complexity in sparse inference.

%%%%%%%%%%%%%%%%%%%%%%%%%%%%%%%%%%%%%%%%%%%%%%%%%%%
%%%%%%%%%%%%%%%%%%%%%%%%%%%%%%%%%%%%%%%%%%%%%%%%%%%
\paragraph{Prior research in neuroscience.}
A key milestone was achieved by \citet{murphy_learning_2012}, who computed non-negative sparse embeddings from word vectors, achieving a significant improvement in interpretability in psychophysical assessments relative to Singular Value Decomposition (SVD). Specifically, they used Non-Negative Sparse Embedding (NNSE), a form of non-negative sparse coding, on word vectors obtained through Positive Pointwise Mutual Information \citep[PPMI,][]{church1990word}. Their approach was tested on three tasks: behavioral experiments (predicting human behavior), neuroscience experiments (predicting fMRI signals), and psychophysics assessments of interpretability. While NNSE and SVD performed similarly on the first two tasks, NNSE notably increased interpretability on the third task, with precision in the interpretability task rising from 46.33\% (for SVD at $k=300$) to 92.33\%.
Building on these results, \citet{fyshe_interpretable_2014} integrated neuroscience data to further refine sparse representations, aligning them with semantic patterns reflected in human brain activity. Such cross-modal constraints, derived from neural data, have also been successfully applied in the visual domain \citep{mcclure_representational_2016,li_learning_2019}.
In our recent work \citep{klindt_identifying_2023}, we observed that virtual neurons—formed as linear combinations of neural activations—can enhance interpretability in artificial networks, on average outperforming individual units in capturing coherent features.
These results were consistent across Convolutional Neural Networks (CNNs) and datasets with neural responses to visual images.

%%%%%%%%%%%%%%%%%%%%%%%%%%%%%%%%%%%%%%%%%%%%%%%%%%%
%%%%%%%%%%%%%%%%%%%%%%%%%%%%%%%%%%%%%%%%%%%%%%%%%%%
\section{A mathematical world model of neural representations}\label{sec:model}
\begin{figure}
    \centering
    \includegraphics[width=0.99\linewidth]{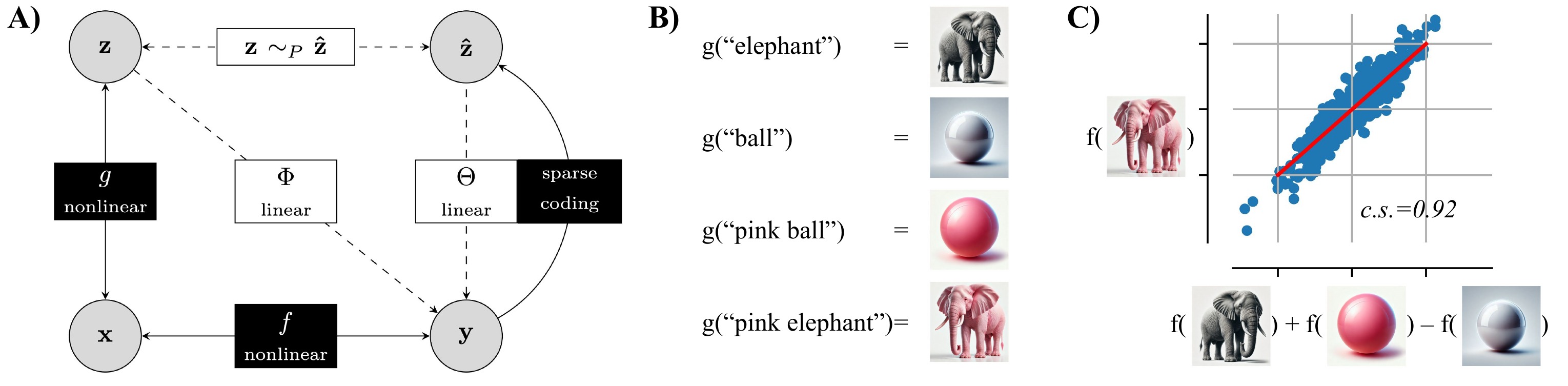}
    \caption{
    \textbf{Mathematical world model and neural analogy making.}
    \textbf{A)}
    % The \textit{latent variables} $z$ can be nonlinearly extracted from data $z=g(x)$.
    The \textit{latent variables} (i.e., \textit{features}, see text) $z$ are nonlinearly mapped to the data $x=g(z)$.
    The data is nonlinearly mapped to neural representations $y=f(x)$.
    % There may be part of the data $\epsilon$ that the latent variables $g(x \times \epsilon) = g(x)$ or the neural representation $f(x \times \epsilon) = f(x)$ are invariant to.
    % However, we do require that, together, $h=f \circ g^{-1}$ are invertible.
    % Under this assumption (and the requirement that all $z$ are important for the task that $f$ is trained to solve), 
    Under the assumption that $h=f \circ g$ is invertible (and the requirement that all $z$ are important for the task that $f$ is trained to solve), 
    the key insight from identifiability theory is that $h(z)=y$ will be linear (Theorem~\ref{thm:ident_theo_supervised}).
    The neural representation is likely a lower-dimensional representation of the, potentially, high-dimensional latent variables, i.e., $h: \mathbb{R}^N \rightarrow \mathbb{R}^M$ with $M \ll N$. 
    Under additional assumptions (e.g., $z$ being sparse) \citep{donoho2006short}, the key insight from compressed sensing theory is that we can perform \textit{sparse coding} to decode sparse codes from the neural representation $y \rightarrow \hat{z}$, allowing us to uniquely identify the true latent variables up to permutations $z \sim_P \hat{z}$.
    \textbf{B)} A generative model (GPT-4) that maps latent variables, here represented as text, to images $g(z)=x$.
    \textbf{C)} A neural representation (ViT-B/16 \citep{dosovitskiy_image_2021}) of the generated images $f(x)$, demonstrating analogy making.
    This shows how concepts combine linearly, at least for this local example, in neural representation space (c.s.: cosine similarity).
    }
    \label{fig:illustration_model}
\end{figure}

We will now establish a mathematical world model to formalize how neural networks encode complex information through superposition. This modeling framework is illustrated in Figs.~\ref{fig:pipeline} (informal) and ~\ref{fig:illustration_model}A (formal).
% We assume that our data $x$ (images, text etc.) contain \textit{latent variables} or \textit{features} $z \in \mathcal{Z}$.
We assume there are underlying \textit{latent variables} $z \in \mathcal{Z}$ responsible for the observed data (we focus on images, but the theory applies equally to other modalities such as language).
These latent variables are likely sparse and high-dimensional, as each image contains only a small subset of all possible objects. To illustrate, consider $z \in \{0, 1\}^N$, where a value of $1$ at index 42 might indicate the presence of an elephant in the scene. 
% We assume the latent variables are independent.
% \charlie{As we will see when we come to superposition, the independence of latent variables is largely what allows representation of a larger number of latent variables than we have dimensions in our dense representations, in whatever form that may be.}
Optionally, we also assume that they are \textit{interpretable}--meaning that a human could predict how changes in $z$ affect the input (we make this notion quantitative in Section~\ref{sec:background_quant_interp}). % , although causal dependencies may complicate this assumption \citep{trauble2021disentangled},
For instance, changing $z_{42}$ from $0$ to $1$ would result in the appearance of an elephant.
Various terms have been used in the literature to refer to $z$, including: \textit{sources} (in ICA) \citep{hyvarinen_identifiability_2023}, \textit{factors of variation} (representation learning) \citep{bengio2013representation}, \textit{latent variables} (probabilistic models) \citep{kingma2013auto,rezende2014stochastic,locatello_challenging_2019} (disentanglement research) \citep{locatello2019challenging}, \textit{features} (neuroscience) \citep{hubel1962receptive,gross1972visual,barlow_single_1972} and (machine learning) \citep{fukushima1969visual} and (mechanistic interpretability) \citep{olah_feature_2017}, or \textit{concepts} (psychology) \citep{quian_invariant_2005}.

Next, we assume a \textit{function} $g: \mathcal{Z} \rightarrow \mathcal{X}$ that maps latent variables to data $x = g(z)$. To illustrate that $g$ is a nonlinear function, consider adding latent variables $z_1 + z_2$, e.g., `elephant' + `pink ball', which we can represent as the conjunction `elephant' $\wedge$ `pink ball' in Smolensky's notation (Equation~\ref{eq:smolensky}). Feeding this combined input into our generator, $g(z_1 + z_2)$, results in an image containing both objects (compare Fig.~\ref{fig:cnn_superposition}). In contrast, adding the individual images $g(z_1)$ and $g(z_2)$ yields a blurry mixed image. Thus, $g(z_1) + g(z_2) \neq g(z_1 + z_2)$, indicating that $g$ must be nonlinear.
It is often assumed that $g$ is invertible \citep{hyvarinen_identifiability_2023}, ensuring that there exists a well-defined map from data $x$ back to latent variables $z$. We refer to $x$ as \textit{data} or \textit{inputs}.

The previous paragraph makes the crucial assumption (identifiablity theory) that the data is generated by randomly drawing latent variables $z$ and passing them through the generative function $g(z)=x$ to obtain the observed data \citep{hyvarinen_independent_2000}.
The assumption of such a generative model or \textit{data generating process} is somewhat controversial, akin to positing the existence of an ideal Platonic reality \citep{huh2024platonic}. 
A much more widely accepted view is that latent variables, or features, can be extracted from the data \citep{elhage_toy_2022}. 
However, the two perspectives are similar enough so that our theory works in both cases:
Specifically, we think of the map $g: \mathcal{Z} \rightarrow \mathcal{X}$ that generates the data and the map $f: \mathcal{X} \rightarrow \mathcal{Y}$ as the neural representation and prove that the combined map $h := f \circ g: \mathcal{Z} \rightarrow \mathcal{Y}$ is linear and invertible.
Alternatively, we may think of the map $g: \mathcal{X} \rightarrow \mathcal{Z}$ that extracts features from the data, but does not explain all of the data in the sense that it is invariant to certain (noise) $\epsilon$ in the data $g(x)=g(\epsilon(x))$; Similarly, the neural representation is invariant to this $f(x)=f(\epsilon(x))$. 
Now, again, all we need is that the combined map $h := f \circ g^{-1}: \mathcal{Z} \rightarrow \mathcal{Y}$ is invertible (for $g$ we have to assume this, for $f$ it follows from reaching the global optimum).
With this construction, our theory works in both frameworks.
However, for this exposition, we will go with the first formalism, because it is easier to explain.

Lastly, we will consider a \textit{representation function} $f: \mathcal{X} \rightarrow \mathcal{Y}$ that corresponds to the hidden activity in a biological or artificial neural network, where $y = f(x)$. 
In most cases, $f$ is also a nonlinear function, which is essential for neural networks to perform complex computations.
Typically, $f$ is a real, vector-valued function with $\mathcal{Y} = \mathbb{R}^M$. While this representation can be high-dimensional---akin to the number of neurons in the visual cortex or channels in a modern LLM---it is most likely still lower than the number of possible latent variables in the world, i.e., $M \ll N$.

%%%%%%%%%%%%%%%%%%%%%%%%%%%%%%%%%%%%%%%%%%%%%%%%%%%
%%%%%%%%%%%%%%%%%%%%%%%%%%%%%%%%%%%%%%%%%%%%%%%%%%%

%%%%%%%%%%%%%%%%%%%%%%%%%%%%%%%%%%%%%%%%%%%%%%%%%%%
%%%%%%%%%%%%%%%%%%%%%%%%%%%%%%%%%%%%%%%%%%%%%%%%%%%
\section{Three steps towards interpretable neural representations}
Building on prior observations that sparse coding extracts interpretable concepts from distributed neural representations (Sec.~\ref{sec:prior_work}) and using our mathematical world model (Sec.~\ref{sec:model}) as a theoretical foundation, we now walk through the three steps of our theory and analysis pipeline (Sec.~\ref{sec:pipeline}).

%%%%%%%%%%%%%%%%%%%%%%%%%%%%%%%%%%%%%%%%%%%%%%%%%%%
%%%%%%%%%%%%%%%%%%%%%%%%%%%%%%%%%%%%%%%%%%%%%%%%%%%
\subsection{Step one: Identifiability theory for linear representations}
Importantly, the true latent variables behind the data, as well as the generating function $g$, are unknown to us. All we observe is the generated data $x$. Our goal is to uncover the true latent variables in the world and understand how our neural networks represent them to perform computations.
To aid in this endeavor, we return to the \textit{superposition principle} \citep{smolensky_tensor_1990, van_gelder_compositionality_1990, elhage_toy_2022}.
Superposition can be nicely observed in \textit{neural analogy-making} \citep{mikolov_efficient_2013}. For example, using textual descriptions for the latent variables like $z = \text{`elephant'}$, consider the operation (see Fig.~\ref{fig:illustration_model}B-C):
\begin{equation}
    f \circ g(\text{`elephant'}) + \textcolor{darkpink}{\underbrace{f \circ g(\text{`pink ball'}) - f \circ g(\text{`ball'})}_{\approx f \circ g(\text{`pink'})}} \approx f \circ g(\text{`pink elephant'})
\end{equation}
with $g(z)=x$ the generative function and $f(x)$ the neural representation.
This implies that there is a direction in representation space corresponding to the concept of `pink', independent of context:
\begin{equation}
    \begin{split}
        \textcolor{darkpink}{f \circ g(\text{`pink'})} &\approx \textcolor{darkpink}{f \circ g(\text{`pink ball'}) - f \circ g(\text{`ball'})} \\
        &\approx \textcolor{darkpink}{f \circ g(\text{`pink elephant'}) - f \circ g(\text{`elephant'})}.
    \end{split}
\end{equation}
The neural representation forms a \textit{superposition} of concepts, such as combining `pink' with `elephant’:
\begin{equation}
    f \circ g(\text{`elephant'}) + \textcolor{darkpink}{f \circ g(`pink')} = f \circ g(\text{`pink elephant'}).
\end{equation}
Datasets such as the Google Analogy Dataset (GA) \citep{mikolov2013efficient} and BATS \citep{drozd2016word} were developed to evaluate neural analogy-making
Generally, the property
\begin{equation}\label{eq:additivity}
    f \circ g(z_1) + f \circ g(z_2) = f \circ g(z_1 + z_2) \quad \forall z_1,z_2 \in \mathcal{Z}
\end{equation}
is referred to as \textit{additivity}.
Together with \textit{homogeneity}, i.e., $f \circ g(\alpha z) = \alpha f \circ g(z), \; \forall z \in \mathcal{Z}, \alpha \in \mathbb{R}$, these are the two defining features of a \textit{linear function}.
In fact, if $f$ and $g$ are continuous (artificial neural networks are usually differentiable), additivity entails homogeneity.
% Thus, assuming that the latent variables live in $\mathcal{Z}=\mathbb{R}^N$, this means that $f \circ g: \mathbb{R}^N \rightarrow \mathbb{R}^M$ is a \textit{linear function}.
Thus, this means that $f \circ g: \mathcal{Z} \rightarrow \mathbb{R}^M$ is a \textit{linear function}.
If $f \circ g$ is linear, this reduces an intractable nonlinear problem \citep{hyvarinen1999nonlinear} into a well-studied linear problem \citep{comon_independent_1994}.
Essentially, it means that the neural representation $f$ inverts the generative model $g$ up to a linear transformation.
% This is a highly nontrivial hypothesis: $g$ and $f$ are nonlinear, but their composition, $f \circ g$, is linear.
These are great news, since we now know that the neural representations are only a linear transformation away from the true latent variables.
The key question is, \textit{why does this happen}?

%%%%%%%%%%%%%%%%%%%%%%%%%%%%%%%%%%%%%%%%%%%%%%%%%%%
%%%%%%%%%%%%%%%%%%%%%%%%%%%%%%%%%%%%%%%%%%%%%%%%%%%
\paragraph{Identifiability up to linear transformations.}\label{sec:ident}
In this section, we briefly present recent results that prove the linear identifiability of a broad class of neural networks trained to do object recognition \citep{reizinger2024cross}.
Prior work proved identifiability of nonlinear ICA in certain probabilistic settings and self-supervised learning scenarios \citep{Hyva16NIPS,morioka_independent_2021,hyvarinen_nonlinear_2017, khemakhem_variational_2020, klindt2020towards,hyvarinen2019nonlinear,khemakhem_ice-beem_2020,morioka_connectivity-contrastive_2023,zimmermann_contrastive_2021}.
As stated by \citet{hyvarinen_nonlinear_2019}, a key part of identifiability theory work, is the assumption of a data-generating process (DGP), or \textit{world model} in the preceding section.
Assuming the cluster-centric DGP in \citep{reizinger2024cross} (\(g: \mathbb{S}^{d-1}\) to \(\mathbb{R}^D\)); we train a continuous encoder $f: \mathbb{R}^{D} \rightarrow \mathbb{R}^{d}$ on $x$, paired with a linear classifier $W$. 
In standard supervised learning, we optimize the cross-entropy objective over all possible encoders $f$ and classifiers $W$.
Now, the key theoretical result:

\vspace{10pt}
\begin{mdframed}
\begin{theorem}[Supervised Learning Identifiability~\citep{reizinger2024cross}]
    \label{thm:ident_theo_supervised}
    Let Assumption 1 in \citet{reizinger2024cross} hold, and suppose that a continuous encoder \(f: \mathbb{R}^D \rightarrow \mathbb{R}^d\) and a linear classifier \(W\) globally minimize the cross-entropy objective. Then, the composition \(h = f \circ g\) is a linear map from \(\mathbb{S}^{d-1}\) to \(\mathbb{R}^d\).
\end{theorem}
\end{mdframed}
\vspace{10pt}

\noindent
Proof provided in \citep{reizinger2024cross}.
The significance of this result is that neural networks trained to classify object categories can learn to invert complex, nonlinear generative models of the world, up to linear transformations.
Despite the complexity of the generative model, Theorem~\ref{thm:ident_theo_supervised} guarantees that we can recover the underlying variables.
This is exciting because it suggests that neural networks may be learning rich representations of the latent variables underlying the world, rather than just memorizing shortcuts to solve a task \citep{lake2017building, geirhos2020shortcut}.
%This contrasts with foundational results by \citet{hyvarinen_nonlinear_1999}, which showed that without further assumptions (e.g., access to class labels), nonlinear ICA is largely intractable. 

%%%%%%%%%%%%%%%%%%%%%%%%%%%%%%%%%%%%%%%%%%%%%%%%%%%
%%%%%%%%%%%%%%%%%%%%%%%%%%%%%%%%%%%%%%%%%%%%%%%%%%%
\paragraph{Limitations and open questions.}
Our framework assumes sparse latent variables in Euclidean space \(\mathbb{R}^n\), whereas contrastive and supervised learning theories often assume dense latent variables on a hypersphere \(\mathbb{S}^{n}\) \citep{zimmermann_contrastive_2021, reizinger2024cross}. Some latent variables, such as object pose, exhibit non-Euclidean topology, requiring embeddings in curved manifolds within the latent space \citep{higgins_towards_2018, pfau_disentangling_2020, keurti_desiderata_2023}. The linear representation hypothesis may hold only approximately in cases where topological constraints prevent a fully linear encoding \citep{smith2024weak_lrh, engels_not_2024}. Extensions of compressed sensing theory suggest that recovery remains feasible even in these cases \citep{baraniuk_random_2009}. Furthermore, we might also assume that latent variables are discrete \citep{park2024geometry}, for which both identifiability \citep{hyttinen2022binary, barin2024identifiability} and compressed sending \citep{stojnic2010recovery} theory exists.
Furthermore, the linear identifiability results (Thm.~\ref{thm:ident_theo_supervised}) assumed that the latent variable dimensionality $\dim(\mathcal{Z})=N$ and that of the neural representation $\dim(\mathcal{Y})=M$ are matched, i.e., $M=N$.
However, a key motivation for looking into superposition and sparse coding was the fact that this is not the case since real-world neural representations often encode many latent variables in a lower-dimensional space ($M<N$) using non-orthogonal directions \citep{elhage_toy_2022}.

%%%%%%%%%%%%%%%%%%%%%%%%%%%%%%%%%%%%%%%%%%%%%%%%%%%
%%%%%%%%%%%%%%%%%%%%%%%%%%%%%%%%%%%%%%%%%%%%%%%%%%%
\subsection{Step two: Compressed sensing and sparse coding}\label{sec:background_compressed_s_sparse_c}
According to the previous section, the superposition principle, which is also the defining feature of a linear function \citep{ill1991}, is equivalent to the \textit{linear representation hypothesis} \citep{arora_latent_2016,park_linear_2024}.
However, \citet{elhage_toy_2022} recently proposed that the geometry of the layout of the neural representations, i.e., non-orthogonality, is a key feature of \textit{superposition}.
If we assume that the latent variables live in $\mathcal{Z}=\mathbb{R}^N$, then this is a necessary consequence because we are linearly projecting $N$ latent variables down into an $M$ dimensional neural representation (although it can also happen when $M \geq N$).
% This is the reason why we need compressed sensing.
To highlight this additional complexity and align with the recent use of the term, we will extend Smolensky's definition (Def.~\ref{def:sup_smol}) of \textit{superposition}:

\begin{mdframed}[nobreak=true]
\begin{definition}[Superposition]\label{def:sup}
    A representation $f$ of the data \(x=g(z)\) with latent variables \(z \in \mathbb{R}^N\) is in \textit{superposition} if \(h=f \circ g\) is a linear function and if there are pairs of vectors that are non-orthogonal \(h(e_i) \not\perp h(e_j)\), where \(e_i\) denote the canonical basis vectors of $\mathbb{R}^N$.
\end{definition}
\end{mdframed}
\vspace{10pt}

\noindent
Thus, a key challenge remains in decoding the sparse (interpretable) concepts from neural representations. For example, the vector $f \circ g(\text{`pink ball'}) - f \circ g(\text{`ball'})$ reveals how the latent variable $z$ ($\text{`pink'}$) is represented, but we are limited by manually defining such pairs.
% However, ideally, we would like an unsupervised way of extracting all the sparse concepts that are represented in superposition in the neural activations.
To `lift' \citep{elhage_toy_2022} these sparse concepts out of superposition, we turn to \textit{compressed sensing} \citep{beurling_sur_1939, donoho2006short, gilbert2002near, candes2006robust, tsaig2006extensions, ganguli_compressed_2012}.
Assuming the setting described above, we know that $f \circ g: \mathbb{R}^N \rightarrow \mathbb{R}^M$ is a linear function.
Thus, there exists a matrix $\Phi \in \mathbb{R}^{M \times N}$ (see Fig.~\ref{fig:illustration_model}A) such that
\begin{equation}
    f \circ g(z) = \Phi z \quad \forall z \in \mathbb{R}^N.
\end{equation}
Moreover, assuming the distribution over latent variables $z \sim P(Z)$ is sparse (at most $K$ active components) \citep{donoho2006short}, we can recover the latent variables with high probability if
\begin{equation}\label{eq:comp_sens}
    M \; > \;  \mathcal{O}\left(K \log \left( \frac{N}{K} \right) \right).
\end{equation}
In compressed sensing we assume knowledge of the linear projection, which is unrealistic, so we have to use \textit{blind compressed sensing} \citep{gleichman2011blind} or \textit{sparse coding} which provides a method to achieve this recovery. It represents data---or, in our setting, neural activations---using a dictionary of basis functions $\Theta \in \mathbb{R}^{M \times N}$ (see Fig.~\ref{fig:illustration_model}A), minimizing both reconstruction error and sparsity. 
% The objective function for a set of neural activations $(y_1, ..., y_n)$, $y^{(i)} \in \mathbb{R}^M$ is defined as:
% \begin{equation}\label{eq:sparse_coding}
%     \mathcal{L}(\Theta, \hat{z}) :=  \sum_{i=1}^D \| y^{(i)} - \Theta \hat{z}_i \|_2^2 + \lambda \|\hat{z}_i\|_1
% \end{equation}
% where the sparse latent variables $\hat{z} \in \mathbb{R}^N$ are typically \textit{overcomplete} ($N > M$).
Sparse coding, for a set of neural activations $(y^{(i)} \in \mathbb{R}^M)_{i=(1,...,D)}$, involves two steps:
\begin{enumerate}
    \item \textbf{Sparse inference:} Given a dictionary $\Theta$, estimate the codes $\hat{z}_i \in \mathbb{R}^N$ for each sample $y^{(i)}$
    \begin{equation}\label{eq:sparse_coding_inf}
        \underset{\hat{z}}{\min} \sum_{i=1}^D \| y^{(i)} - \Theta \hat{z}_i \|_2^2 + \lambda \|\hat{z}_i\|_1.
    \end{equation}
    \item \textbf{Dictionary learning}: Given codes $\hat{z}$, update the dictionary $\Theta$ (fixing norms to avoid collapse)
    \begin{equation}\label{eq:sparse_coding_dict}
        \underset{\Theta}{\min} \sum_{i=1}^D \| y^{(i)} - \Theta \hat{z}_i \|_2^2 \quad \text{s.t.} \quad \forall j \in \{1, ..., N\}: \|\Theta_{:,j}\|=1.
    \end{equation}
\end{enumerate}
In summary, by applying sparse coding to neural representations, we extract sparse, interpretable codes that can `lift' neural representations out of superposition \citep{yun_transformer_2021}. 
That is, if $y=f \circ g(z)=\Phi z$ is in \textit{superposition} (Def.~\ref{def:sup}), then sparse inference should yield back the true latent variables $\hat{z} \sim z$ up to permutations and scaling \citep{hillar2015short}.
This approach allows us to map neural representations back to human-interpretable concepts \citep{arora_linear_2018}.

The difficulty with this method is that sparse inference is non-convex (and NP-hard) \citep{natarajan1995sparse}, that means there are many possible $\hat{z}$ that \textit{locally} minimize \ref{eq:sparse_coding_inf}.
Moreover, for a large dataset ($D$) this is costly, because (1.) has to be run for every dictionary training step and (2.) for every sample at inference time.
Consequently, many sparse coding algorithms have been proposed over the years as can be seen from the sample of methods in Tab.~\ref{tab:prior_results}.
Next, we will discuss one of the most popular recent methods, i.e., \textit{sparse autoencoders} \citep{ng2011sparse, bricken_towards_2023}.

\paragraph{Sparse autoencoders.}\label{sec:sparse_coding}
% Sparse coding involves representing data using a small number of active units, making the representations more interpretable.
To run sparse inference (\ref{eq:sparse_coding_inf}), for computational efficiency, especially with large datasets, one can employ \textit{amortized inference} \citep{kingma2013auto,rezende2014stochastic}. Instead of solving the optimization problem for each $y^{(i)}$, we learn a function $\xi$ that maps inputs directly to their sparse codes  $\hat{z_i} = \xi(y^{(i)})$ \citep{vafaii_poisson_2024}.
When this mapping takes a simple linear-nonlinear form, $\xi(y^{(i)}) = \text{ReLU}(W y^{(i)} + \beta)$, this is called a \textit{sparse autoencoder} (SAE) \citep{ng2011sparse, bricken_towards_2023}, whose computational properties we discuss theoretically below.
We will consider setting where sparse signals $z \in \mathbb{R}^N$ with at most $K$ non-zero entries are projected into an $M$-dimensional space ($M < N$). Compressed sensing theory guarantees that these signals can be uniquely recovered with sparse coding \citep{donoho2006short}. However, recent results proved that SAEs fail to achieve optimal recovery \citep{oneill2025computeoptimalinferenceprovable}.
Intuitively, this limitation stems from the architecture of the SAE’s linear-nonlinear encoder, which lacks the necessary complexity to fully recover the high-dimensional ($N$) sparse representation from the lower-dimensional ($M$) projection.
Equipping SAEs with more expressive (MLP) encoders increases code recovery and interpretability \citep{oneill2025computeoptimalinferenceprovable}.
Suboptimal sparse coding models (such as SAEs) with optimization and inference issues might explain why, empirically, the extracted features are not always unique and universal across models \citep{paulo2025sparse}, despite theory \citep{hillar_when_2015,olshausen08sc_ica,aharon_uniqueness_2006, hillar2015short}.

%%%%%%%%%%%%%%%%%%%%%%%%%%%%%%%%%%%%%%%%%%%%%%%%%%%%%%%%%%%%
%%%%%%%%%%%%%%%%%%%%%%%%%%%%%%%%%%%%%%%%%%%%%%%%%%%%%%%%%%%%

\begin{figure}[tb]
    \centering
    \includegraphics[width=0.99\textwidth]{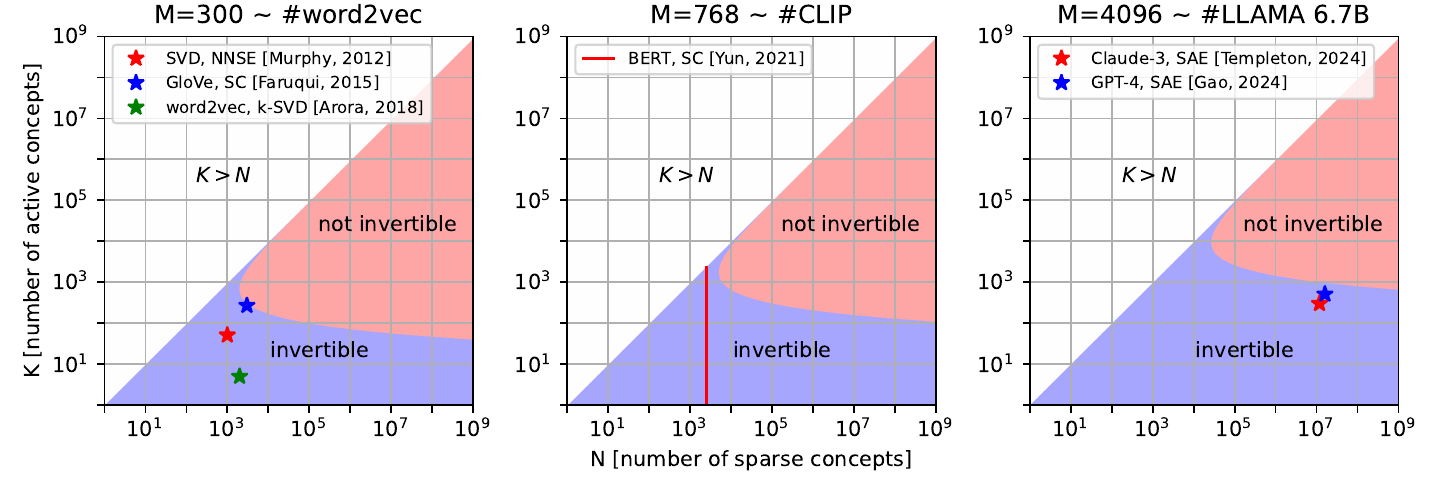}
    \caption{
    \textbf{Compressed sensing bounds.}
    For various embedding sizes ($M$), the plots show lower bounds for signal reconstruction with high probability (omitting any constant scaling factors) \citep{donoho2006short}. The boundary between invertible and non-invertible regions is defined by $M = K \log(N/K)$. Shown are embeddings for hidden dimension sizes of models such as \texttt{word2vec} \citep{mikolov_efficient_2013}, \texttt{CLIP} \citep{radford_learning_2021}, and \texttt{LLAMAv1} with 6.7B parameters \citep{touvron_llama_2023}.
    }
    \label{fig:CS_rec}
    \vspace{0pt}
\end{figure}

%%%%%%%%%%%%%%%%%%%%%%%%%%%%%%%%%%%%%%%%%%%%%%%%%%%
%%%%%%%%%%%%%%%%%%%%%%%%%%%%%%%%%%%%%%%%%%%%%%%%%%%
\paragraph{Limitations and open questions.}
Recent investigations have scaled up SAEs significantly---Fig.~\ref{fig:CS_rec} shows the conditions under which compressed sensing theory (\ref{eq:comp_sens}) predicts that codes can be recovered. Our results indicate that simple SAEs cannot achieve optimal recovery, raising the question of what types of sparse coding algorithms can scale to handle tens of millions of sparse codes \citep{templeton_scaling_2024, gao_scaling_2024}. Answering this and designing efficient sparse inference algorithms that close the amortization gap is crucial for further progress in decoding high-dimensional neural representations. 
It is also an open question of how sparse coding methods perform when the input to the model is shifted away from the training data \citep{paiton2020selectivity,friedman2023interpretability}.
Furthermore, \citet{wright2024addressing} identified feature suppression in SAEs, which is equivalent to the activation shrinkage first described by \citet{tibshirani_regression_1996} as a property of L1 penalties. Numerous SAE architectures have since been proposed to address this, from gated SAEs \citep{rajamanoharan_improving_2024} to JumpReLU SAEs \citep{rajamanoharan2024jumpingaheadimprovingreconstruction}.

%%%%%%%%%%%%%%%%%%%%%%%%%%%%%%%%%%%%%%%%%%%%%%%%%%%
%%%%%%%%%%%%%%%%%%%%%%%%%%%%%%%%%%%%%%%%%%%%%%%%%%%
\subsection{Step three: Quantitative interpretability for evaluation}
To validate our ability to recover interpretable features up to permutation, we need a quantitative evaluation metric.
% Now that we have a theory and an analysis pipeline that suggests we are able to recover interpretable features up to permutations, we need a way to verify this.
One approach, common in nonlinear ICA literature, is to simulate data and see how well we recover the true (simulated) features.
This is not viable on real data.
However, we can make a simple assumption: \textit{the true features are more interpretable than linear combinations of them}.
% For instance, a neuron that fires only when an elephant is an image, is more interpretable than a neuron that also fires when a ball is in an image.
For instance, a neuron selective for elephants is more interpretable than one responding to both elephants and balls.
Formally, assume we have a function to measure interpretability $I: \mathcal{F} \to [0, 1]$ that takes a neuron (i.e., a function on inputs $f_i: \mathcal{X} \to \mathbb{R}$) and returns a value between zero (uninterpretable) and one (highly interpretable).
The interpretability of a population code, e.g., with $M$ neurons is simply the average $I(f) := \frac{1}{M}\sum_i^M I(f_i)$.
% For instance, the function that extracts the latent variables, i.e., interpretable features $g^{-1}: \mathcal{X} \to \mathbb{R}^N$ would score high by definition $I(g^{-1}) = 1$.
By definition, the function extracting latent variables, $g^{-1}: \mathcal{X} \to \mathbb{R}^n$, is maximally interpretable: $I(g^{-1}) = 1$.
Now, saying that combinations of features are less interpretable means, for any two distinct $i \neq j$ neurons $g^{-1}_i, g^{-1}_j$ and $a, b \in \mathbb{R}$
\begin{equation}
    I(a g^{-1}_i + b g^{-1}_j) < 1.
\end{equation}
% Consequently, for almost all linear transformations $\Theta: \mathbb{R}^M \to \mathbb{R}^M$, we have $I(\Theta f) < I(f)$.
Since most linear transformations (except permutations) \textit{mix} feature representations, almost all $\Theta$ reduce interpretability: $I(\Theta g^{-1}) < I(g^{-1})$.
Especially in superposition with $M<N$ this becomes worse as we are mixing even more concepts into each neuron.
Since, $f = \Theta g^{-1}$, that means $I(f) < I(g^{-1})$, i.e., the neurons in our model are less interpretable than the latent variables.
The only exception are permutations $P \in S_n$ because the sum is invariant to permutations
\begin{equation}
    I(P g^{-1}) = \frac{1}{M}\sum_i^M I(P_i g^{-1}) = \frac{1}{M}\sum_i^M I(g^{-1}_{\pi(i)}) = \frac{1}{M}\sum_{j=\pi(i)}^M I(g^{-1}_j) = I(g^{-1})
\end{equation}
where $\pi(i)=j$ is the single non-zero entry in the i-th row of $P$.
In summary, an interpretability metric invariant to permutations would allow us to assess how well our extracted sparse codes recover the true underlying features. Next, we will discuss such a metric.

%%%%%%%%%%%%%%%%%%%%%%%%%%%%%%%%%%%%%%%%%%%%%%%%%%%
%%%%%%%%%%%%%%%%%%%%%%%%%%%%%%%%%%%%%%%%%%%%%%%%%%%
\paragraph{Quantitative interpretability methods.}\label{sec:background_quant_interp}
Quantitative assessment of interpretability is crucial to move beyond anecdotal analyses \citep{doshi-velez_towards_2017,leavitt_towards_2020}. 
One approach is to use psychophysics tasks, which assess human understanding of neural network computations \citep{berardino_eigen-distortions_2018}. 
These tasks operationalize \textit{interpretability} by testing whether humans can perform tasks that require understanding the model’s output.
Understanding is a difficult philosophical concept and yet, to test whether a person has understood something, we can test them on a \textit{behavioral} task that requires understanding \citep{wittgenstein_philosophische_1953}.
For \textbf{language models}, human understanding of neural representations can be tested, for instance, through the \textit{word intrusion task} (WIT), where participants identify an outlier word from a list of related terms. For example, in the set \textit{\{one, two, three, elephant\}}, \textit{elephant} is the intruder. 
% Formally, we compute a unit’s activations $f_j(X)$ over the dataset and define a set $C(f_j)$ of highly activating examples. The task is to identify the outlier in $C(f_j) \cup \{x_{n}\}$ \citep{chang_reading_2009}. 
This approach is analogous to the \textit{New York Times Connections Game}.
A related task is the \textit{police lineup task} \citep{arora_linear_2018}, where participants identify the true senses of polysemous words from distractors. Each sense is represented by a set of related words, and participants must select the correct sense. This task quantitatively assesses interpretability by comparing human performance to model predictions, using WordNet’s database of polysemous words \citep{poli_wordnet_2010}.
Beyond representation, we also want to evaluate whether \textit{inference} yields interpretable results. In the \textit{topic intrusion task} (TI) \citep{chang_reading_2009}, the challenge is to associate input $x$ with the correct unit in the network, based on the population response $f(x)$. The task measures whether the inferred codes are interpretable across various inputs. 
While WIT tests \textit{specificity}, the TI tasks might be a good operationalization to test \textit{sensitivity}, which has proven to be harder to assess \cite{templeton_scaling_2024}.
For \textbf{vision models}, human understanding of neural representations can be tested, for instance, by adapting the \textit{intrusion task} \citep{ghorbani2019towards} or through the two-alternative forced choice task proposed by \citet{borowski_exemplary_2021}. In this task, participants match query images to visual explanations based on unit activations. Our recent work automates this process \citep{klindt_identifying_2023}), allowing faster evaluations while maintaining strong correlation with human judgments. 
This allows scaling vision model interpretability analyses to more than $70$ million units from $835$ computer vision models \citep{zimmermann_measuring_2024}.

% \begin{figure}[htb!]
%     \centering
%     \includegraphics[width=0.9\linewidth]{figures/fig_roland_mi.pdf}
%     \caption{
%         \textbf{Machine Interpretability Score (MIS) Framework.}
%         \textbf{A.} Psychophysics task for measuring interpretability \citep{borowski_exemplary_2021}: participants match query images to references based on activation. Red and blue squares indicate images that minimally and maximally activate a neuron.
%         \textbf{B.} MIS automates this process \citep{klindt_identifying_2023, zimmermann_measuring_2024}, using a feature encoder and binary classifier to compute image similarities.
%         \textbf{C.} MIS is highly correlated with human ratings, enabling efficient interpretability quantification.
%     }
%     \label{fig:mis_schematics}
% \end{figure}

%%%%%%%%%%%%%%%%%%%%%%%%%%%%%%%%%%%%%%%%%%%%%%%%%%%
%%%%%%%%%%%%%%%%%%%%%%%%%%%%%%%%%%%%%%%%%%%%%%%%%%%
\paragraph{Limitations and open questions.}
At a high level, all of those behavioral assays depend on a human (dis)similarity judgment between two inputs $d_h: \mathcal{X} \times \mathcal{X} \to \mathbb{R}$ from a dataset $(x_1, ..., x_D)$.
If $d_h$ were a simple, context-free function, we could conceptualize the problem of human interpretability tests in the following way.
We can also define a distance for a scalar representation function $f: \mathcal{X} \rightarrow \mathbb{R}$ like $d_f(x_i, x_j):=|f(x_i) - f(x_j)|$.
The most interpretable representation maintains the ordering of the data (in human similarity judgments) as much as possible, i.e.,
\begin{equation}
    f^* = \underset{f}{\text{argmin}} \sum_{i,j}^D |d_h(x_i, x_j) - d_f(x_i, x_j) |
\end{equation}
Note, that this can be seen as the stress function in multidimensional scaling (MDS)---with a one dimensional embedding \citep{mead1992review,borg2007modern}.
However, the big problem is that $d_h$ depends on context $c$ as the following example demonstrates.
Consider the word intrusion task \textit{\{apple, orange, android\}}.
One possible interpretation is that this is about \textit{fruits}, implying that $d_h(apple, orange|c=fruits) < d_h(apple, android|c=fruits)$ (same for orange/apple and orange/android) 
and, thus, \textit{android} is the intruder.
A different possible interpretation is that this is about operating systems (\textit{OS}), implying that $d_h(apple, android|c=OS) < d_h(apple, orange|c=OS)$ (same for android/apple and android/orange)
and, thus, \textit{orange} would be the intruder.
Presumably, in the word intrusion task \citep{chang_reading_2009} but also in \citet{borowski_exemplary_2021}'s visual psychophysics task, humans are able to infer the context from all provided references.
This illustrates that interpretability scores based on pairwise human (dis)similarity judgments are insufficient and motivates future research into contextualized similarity metrics.

%%%%%%%%%%%%%%%%%%%%%%%%%%%%%%%%%%%%%%%%%%%%%%%%%%%
%%%%%%%%%%%%%%%%%%%%%%%%%%%%%%%%%%%%%%%%%%%%%%%%%%%
\section{Conclusion}

In this perspective paper, we explored the remarkable emergence of \textit{linear representations} in neural networks, despite their inherently nonlinear architectures. We introduced a theoretical framework suggesting that neural networks encode information through \textit{superposition}, in which multiple concepts are linearly and non-orthogonally combined within their representations. Recent theoretical results (Theorem~\ref{thm:ident_theo_supervised}, \citep{hyvarinen_nonlinear_2019,zimmermann_contrastive_2021,reizinger2024cross}) show that neural networks trained on classification tasks can learn to invert complex, nonlinear generative models of the world—up to a linear transformation. By drawing connections to compressed sensing \citep{donoho2006short,baraniuk_random_2009} and sparse coding \citep{olshausen08sc_ica,aharon_uniqueness_2006}, we argued that interpretable, sparse codes can be recovered from superposed neural activations. In addition, we highlighted the importance of developing quantitative measures of interpretability through both theoretical frameworks and behavioral tasks \citep{chang_reading_2009,borowski_exemplary_2021,klindt_identifying_2023} to systematically assess the recovered features.

These findings carry significant implications for theories of neural coding and AI transparency. In neuroscience, the longstanding debate between the single neuron doctrine—where individual neurons are thought to encode discrete features \citep{barlow_single_1972,thorpe_local_1989,hubel1962receptive}—and population coding is re-examined in light of our perspective. While some neurons exhibit \textit{clean selectivity} / \textit{monosemanticity} \citep{rigotti_importance_2013,bricken2022monosemanticity}, other studies have shown that meaningful representations arise from distributed, population-level codes, where neurons display \textit{mixed selectivity} / \textit{polysemanticity} \citep{klindt_identifying_2023}. Future work should aim to explain why, for instance, population codes are prevalent in motor control \citep{mazor2005transient} but appear less frequently in visual areas \citep{saxena2019towards}, suggesting that the brain’s coding principles may flexibly balance sparse, \textit{privileged axes} \citep{khosla_privileged_2024} with distributed representations. At the same time, our work has implications for AI transparency, as it grounds interpretability in rigorous theoretical guarantees via nonlinear ICA and sparse coding, we can build models that not only perform well but also offer insights into the underlying joint distributions governing complex natural phenomena \citep{klindt2024towards,simon2024interplm,schuster2024can}.

Despite these advances, several important open questions remain. Among the many problems with distributed representations \citep{fodor_connectionism_1988}, one key challenge is the \textit{binding problem} \citep{smolensky_tensor_1990,treisman1996binding,feng2023language,feng2024monitoring}-—how multiple objects and attributes (e.g., ``a blue triangle and a red square'') are encoded without simply summing individual representation vectors which would result in a loss of relational information: \(f\circ g(\text{``blue''}) + f\circ g(\text{``triangle''}) + f\circ g(\text{``red''}) + f\circ g(\text{``square''})\) is the same as \(f\circ g(\text{``blue''}) + f\circ g(\text{``square''}) + f\circ g(\text{``red''}) + f\circ g(\text{``triangle''})\). Moreover, while compressed sensing theory provides guarantees for sparse recovery under ideal conditions, practical recovery in large-scale models remains challenging, necessitating the development of more efficient and scalable sparse coding algorithms \citep{oneill2025computeoptimalinferenceprovable}. Extending our framework to incorporate latent variables with non-Euclidean geometry or discrete structure \citep{park2024geometry,hyttinen2022binary} is also critical for a more complete understanding.

Finally, the connections we have drawn between (nonlinear) ICA and sparse coding are bolstered by both classical work \citep{olshausen08sc_ica,aharon_uniqueness_2006,hillar2015short} and recent evidence showing that sparse codes extracted from different networks converge on similar concepts \citep{lan_sparse_2024,chen2023canonical}. Advances in nonlinear ICA and causal inference increasingly rely on sparsity assumptions for identifiability \citep{klindt2021towards,lachapelle_synergies_2023,xu_sparsity_2024,zheng_generalizing_2023}, yet extending these results to cover scenarios involving lower-dimensional latent spaces, manifolds, and anisotropy \cite{rusak_infonce_2024} remains an open and exciting challenge. In summary, by integrating insights from theoretical neuroscience, representation learning, and interpretability research, our work lays a solid foundation for future studies aimed at deciphering the coding principles of both artificial and biological neural systems, while also advancing the development of transparent and robust AI models.

%%%%%%%%%%%%%%%%%%%%%%%%%%%%%%%%%%%%%%%%%%%%%%%%%%%
%%%%%%%%%%%%%%%%%%%%%%%%%%%%%%%%%%%%%%%%%%%%%%%%%%%
\section*{Acknowledgments}
We would like to thank Sung Ching Liu, Hermanni H\"alv\"a, Aapo Hyv\"arinen, Habon Issa, Richard Baraniuk, Zeyu Yun, Bruno Olshausen, Roland Zimmermann, Wieland Brendel and Christian Intern\`o for their feedback, suggestions and pointers that helped us in writing this manuscript.
Moreover, we thank Trevor Christensen, Alice Rigg, Jake Ward, Josh Engels, Katrin Franke, Nancy Kanwisher, Andreas Tolias, Sophia Sanborn, Xaq Pitkow, Nikos Karantzas, Liam Storan, Randall Balestriero, Alice Bizeul, Attila Juhos, Mark Ibrahim, Daniel Kunin, Cl\'ementine Domin\'e, Marco Fumero and Feng Chen for helpful discussions on this topic.

%%%%%%%%%%%%%%%%%%%%%%%%%%%%%%%%%%%%%%%%%%%%%%%%%%%
%%%%%%%%%%%%%%%%%%%%%%%%%%%%%%%%%%%%%%%%%%%%%%%%%%%
% \newpage
\setcitestyle{numbers} % set the citation style to ``numbers''.
\bibliography{references}

\begin{thebibliography}{151}
\providecommand{\natexlab}[1]{#1}
\providecommand{\url}[1]{\texttt{#1}}
\expandafter\ifx\csname urlstyle\endcsname\relax
  \providecommand{\doi}[1]{doi: #1}\else
  \providecommand{\doi}{doi: \begingroup \urlstyle{rm}\Url}\fi

\bibitem[Aharon et~al.(2006)Aharon, Elad, and Bruckstein]{aharon_uniqueness_2006}
Michal Aharon, Michael Elad, and Alfred~M. Bruckstein.
\newblock On the uniqueness of overcomplete dictionaries, and a practical way to retrieve them.
\newblock \emph{Linear Algebra and its Applications}, 416\penalty0 (1):\penalty0 48--67, jul 2006.
\newblock ISSN 0024-3795.
\newblock \doi{10.1016/j.laa.2005.06.035}.
\newblock URL \url{https://www.sciencedirect.com/science/article/pii/S0024379505003459}.

\bibitem[Arora et~al.(2016)Arora, Li, Liang, Ma, and Risteski]{arora_latent_2016}
Sanjeev Arora, Yuanzhi Li, Yingyu Liang, Tengyu Ma, and Andrej Risteski.
\newblock A {Latent} {Variable} {Model} {Approach} to {PMI}-based {Word} {Embeddings}.
\newblock \emph{Transactions of the Association for Computational Linguistics}, 4:\penalty0 385--399, dec 2016.
\newblock ISSN 2307-387X.
\newblock \doi{10.1162/tacl_a_00106}.
\newblock URL \url{https://direct.mit.edu/tacl/article/43373}.

\bibitem[Arora et~al.(2018)Arora, Li, Liang, Ma, and Risteski]{arora_linear_2018}
Sanjeev Arora, Yuanzhi Li, Yingyu Liang, Tengyu Ma, and Andrej Risteski.
\newblock Linear {Algebraic} {Structure} of {Word} {Senses}, with {Applications} to {Polysemy}.
\newblock \emph{Transactions of the Association for Computational Linguistics}, 6:\penalty0 483--495, dec 2018.
\newblock ISSN 2307-387X.
\newblock \doi{10.1162/tacl_a_00034}.
\newblock URL \url{https://direct.mit.edu/tacl/article/43451}.

\bibitem[Baraniuk and Wakin(2009)]{baraniuk_random_2009}
Richard~G. Baraniuk and Michael~B. Wakin.
\newblock Random {Projections} of {Smooth} {Manifolds}.
\newblock \emph{Foundations of Computational Mathematics}, 9\penalty0 (1):\penalty0 51--77, feb 2009.
\newblock ISSN 1615-3375, 1615-3383.
\newblock \doi{10.1007/s10208-007-9011-z}.
\newblock URL \url{http://link.springer.com/10.1007/s10208-007-9011-z}.

\bibitem[Barin-Pacela et~al.(2024)Barin-Pacela, Ahuja, Lacoste-Julien, and Vincent]{barin2024identifiability}
Vit{\'o}ria Barin-Pacela, Kartik Ahuja, Simon Lacoste-Julien, and Pascal Vincent.
\newblock On the identifiability of quantized factors.
\newblock In \emph{Causal Learning and Reasoning}, pages 384--422. PMLR, 2024.

\bibitem[Barlow(1972)]{barlow_single_1972}
H.~B. Barlow.
\newblock Single units and sensation: a neuron doctrine for perceptual psychology?
\newblock \emph{Perception}, 1\penalty0 (4):\penalty0 371--394, 1972.
\newblock ISSN 0301-0066.
\newblock \doi{10.1068/p010371}.

\bibitem[Bengio et~al.(2013)Bengio, Courville, and Vincent]{bengio2013representation}
Yoshua Bengio, Aaron Courville, and Pascal Vincent.
\newblock Representation learning: A review and new perspectives.
\newblock \emph{IEEE transactions on pattern analysis and machine intelligence}, 35\penalty0 (8):\penalty0 1798--1828, 2013.

\bibitem[Berardino et~al.(2018)Berardino, Ballé, Laparra, and Simoncelli]{berardino_eigen-distortions_2018}
Alexander Berardino, Johannes Ballé, Valero Laparra, and Eero~P. Simoncelli.
\newblock Eigen-{Distortions} of {Hierarchical} {Representations}, feb 2018.
\newblock URL \url{http://arxiv.org/abs/1710.02266}.
\newblock arXiv:1710.02266 [cs].

\bibitem[Beurling(1939)]{beurling_sur_1939}
Arne Beurling.
\newblock Sur les intégrales de {Fourier} absolument convergents et leur application à une transformation functionelle, 1939.

\bibitem[Blei et~al.(2003)Blei, Ng, and Jordan]{blei_latent_2003}
David~M. Blei, Andrew~Y. Ng, and Michael~I. Jordan.
\newblock Latent dirichlet allocation.
\newblock \emph{Journal of machine Learning research}, 3\penalty0 (Jan):\penalty0 993--1022, 2003.
\newblock URL \url{https://www.jmlr.org/papers/volume3/blei03a/blei03a.pdf?ref=http://githubhelp.com}.

\bibitem[Borg and Groenen(2007)]{borg2007modern}
Ingwer Borg and Patrick~JF Groenen.
\newblock \emph{Modern multidimensional scaling: Theory and applications}.
\newblock Springer Science \& Business Media, 2007.

\bibitem[Borowski et~al.(2021)Borowski, Zimmermann, Schepers, Geirhos, Wallis, Bethge, and Brendel]{borowski_exemplary_2021}
Judy Borowski, Roland~S. Zimmermann, Judith Schepers, Robert Geirhos, Thomas S.~A. Wallis, Matthias Bethge, and Wieland Brendel.
\newblock Exemplary {Natural} {Images} {Explain} {CNN} {Activations} {Better} than {State}-of-the-{Art} {Feature} {Visualization}, may 2021.
\newblock URL \url{http://arxiv.org/abs/2010.12606}.
\newblock arXiv:2010.12606 [cs].

\bibitem[Bricken et~al.(2023{\natexlab{a}})Bricken, Templeton, Batson, Chen, Jermyn, Conerly, Turner, Anil, Denison, and Askell]{bricken_towards_2023}
Trenton Bricken, Adly Templeton, Joshua Batson, Brian Chen, Adam Jermyn, Tom Conerly, Nick Turner, Cem Anil, Carson Denison, and Amanda Askell.
\newblock Towards monosemanticity: {Decomposing} language models with dictionary learning.
\newblock \emph{Transformer Circuits Thread}, page~2, 2023{\natexlab{a}}.
\newblock URL \url{https://transformer-circuits.pub/2023/monosemantic-features}.

\bibitem[Bricken et~al.(2023{\natexlab{b}})Bricken, Templeton, Batson, Chen, Jermyn, Conerly, Turner, Anil, Denison, Askell, Lasenby, Wu, Kravec, Schiefer, Maxwell, Joseph, Hatfield-Dodds, Tamkin, Nguyen, McLean, Burke, Hume, Carter, Henighan, and Olah]{bricken2022monosemanticity}
Trenton Bricken, Adly Templeton, Joshua Batson, Brian Chen, Adam Jermyn, Tom Conerly, Nick Turner, Cem Anil, Carson Denison, Amanda Askell, Robert Lasenby, Yifan Wu, Shauna Kravec, Nicholas Schiefer, Tim Maxwell, Nicholas Joseph, Zac Hatfield-Dodds, Alex Tamkin, Karina Nguyen, Brayden McLean, Josiah~E Burke, Tristan Hume, Shan Carter, Tom Henighan, and Christopher Olah.
\newblock Towards monosemanticity: Decomposing language models with dictionary learning.
\newblock \emph{Transformer Circuits Thread}, 2023{\natexlab{b}}.
\newblock https://transformer-circuits.pub/2023/monosemantic-features/index.html.

\bibitem[Cand{\`e}s et~al.(2006)Cand{\`e}s, Romberg, and Tao]{candes2006robust}
Emmanuel~J Cand{\`e}s, Justin Romberg, and Terence Tao.
\newblock Robust uncertainty principles: Exact signal reconstruction from highly incomplete frequency information.
\newblock \emph{IEEE Transactions on information theory}, 52\penalty0 (2):\penalty0 489--509, 2006.

\bibitem[Chang et~al.(2009)Chang, Gerrish, Wang, Boyd-graber, and Blei]{chang_reading_2009}
Jonathan Chang, Sean Gerrish, Chong Wang, Jordan Boyd-graber, and David Blei.
\newblock Reading {Tea} {Leaves}: {How} {Humans} {Interpret} {Topic} {Models}.
\newblock In \emph{Advances in {Neural} {Information} {Processing} {Systems}}, volume~22. Curran Associates, Inc., 2009.

\bibitem[Chen and Bonner(2023)]{chen2023canonical}
Zirui Chen and Michael Bonner.
\newblock Canonical dimensions of neural visual representation.
\newblock \emph{Journal of Vision}, 23\penalty0 (9):\penalty0 4937--4937, 2023.

\bibitem[Chomsky(2014)]{chomsky2014minimalist}
Noam Chomsky.
\newblock \emph{The minimalist program}.
\newblock MIT press, 2014.

\bibitem[Church and Hanks(1990)]{church1990word}
Kenneth Church and Patrick Hanks.
\newblock Word association norms, mutual information, and lexicography.
\newblock \emph{Computational linguistics}, 16\penalty0 (1):\penalty0 22--29, 1990.

\bibitem[Comon(1994)]{comon_independent_1994}
Pierre Comon.
\newblock Independent component analysis, a new concept?
\newblock \emph{Signal processing}, 36\penalty0 (3):\penalty0 287--314, 1994.
\newblock URL \url{https://www.sciencedirect.com/science/article/pii/0165168494900299}.
\newblock Publisher: Elsevier.

\bibitem[Devlin et~al.(2019)Devlin, Chang, Lee, and Toutanova]{devlin_bert_2019}
Jacob Devlin, Ming-Wei Chang, Kenton Lee, and Kristina Toutanova.
\newblock {BERT}: {Pre}-training of {Deep} {Bidirectional} {Transformers} for {Language} {Understanding}, may 2019.
\newblock URL \url{http://arxiv.org/abs/1810.04805}.
\newblock arXiv:1810.04805 [cs].

\bibitem[Donoho(2006)]{donoho2006short}
David~L. Donoho.
\newblock Compressed sensing.
\newblock \emph{IEEE Transactions on information theory}, 52\penalty0 (4):\penalty0 1289--1306, 2006.
\newblock Publisher: IEEE.

\bibitem[Doshi-Velez and Kim(2017)]{doshi-velez_towards_2017}
Finale Doshi-Velez and Been Kim.
\newblock Towards {A} {Rigorous} {Science} of {Interpretable} {Machine} {Learning}, mar 2017.
\newblock URL \url{http://arxiv.org/abs/1702.08608}.
\newblock arXiv:1702.08608 [cs, stat].

\bibitem[Dosovitskiy et~al.(2021)Dosovitskiy, Beyer, Kolesnikov, Weissenborn, Zhai, Unterthiner, Dehghani, Minderer, Heigold, Gelly, Uszkoreit, and Houlsby]{dosovitskiy_image_2021}
Alexey Dosovitskiy, Lucas Beyer, Alexander Kolesnikov, Dirk Weissenborn, Xiaohua Zhai, Thomas Unterthiner, Mostafa Dehghani, Matthias Minderer, Georg Heigold, Sylvain Gelly, Jakob Uszkoreit, and Neil Houlsby.
\newblock An {Image} is {Worth} 16x16 {Words}: {Transformers} for {Image} {Recognition} at {Scale}, jun 2021.
\newblock URL \url{http://arxiv.org/abs/2010.11929}.
\newblock arXiv:2010.11929 [cs].

\bibitem[Drozd et~al.(2016)Drozd, Gladkova, and Matsuoka]{drozd2016word}
Aleksandr Drozd, Anna Gladkova, and Satoshi Matsuoka.
\newblock Word embeddings, analogies, and machine learning: Beyond king-man+ woman= queen.
\newblock In \emph{Proceedings of coling 2016, the 26th international conference on computational linguistics: Technical papers}, pages 3519--3530, 2016.

\bibitem[Elhage et~al.(2022)Elhage, Hume, Olsson, Schiefer, Henighan, Kravec, Hatfield-Dodds, Lasenby, Drain, Chen, Grosse, McCandlish, Kaplan, Amodei, Wattenberg, and Olah]{elhage_toy_2022}
Nelson Elhage, Tristan Hume, Catherine Olsson, Nicholas Schiefer, Tom Henighan, Shauna Kravec, Zac Hatfield-Dodds, Robert Lasenby, Dawn Drain, Carol Chen, Roger Grosse, Sam McCandlish, Jared Kaplan, Dario Amodei, Martin Wattenberg, and Christopher Olah.
\newblock Toy {Models} of {Superposition}, sep 2022.
\newblock URL \url{http://arxiv.org/abs/2209.10652}.
\newblock arXiv:2209.10652 [cs].

\bibitem[Engels et~al.(2024)Engels, Liao, Michaud, Gurnee, and Tegmark]{engels_not_2024}
Joshua Engels, Isaac Liao, Eric~J. Michaud, Wes Gurnee, and Max Tegmark.
\newblock Not {All} {Language} {Model} {Features} {Are} {Linear}, may 2024.
\newblock URL \url{http://arxiv.org/abs/2405.14860}.
\newblock arXiv:2405.14860 [cs].

\bibitem[Fan et~al.(2021)Fan, Xiong, Li, and Wang]{fan2021interpretability}
Feng-Lei Fan, Jinjun Xiong, Mengzhou Li, and Ge~Wang.
\newblock On interpretability of artificial neural networks: A survey.
\newblock \emph{IEEE Transactions on Radiation and Plasma Medical Sciences}, 5\penalty0 (6):\penalty0 741--760, 2021.

\bibitem[Faruqui et~al.(2015)Faruqui, Tsvetkov, Yogatama, Dyer, and Smith]{faruqui_sparse_2015}
Manaal Faruqui, Yulia Tsvetkov, Dani Yogatama, Chris Dyer, and Noah Smith.
\newblock Sparse {Overcomplete} {Word} {Vector} {Representations}, jun 2015.
\newblock URL \url{http://arxiv.org/abs/1506.02004}.
\newblock arXiv:1506.02004 [cs].

\bibitem[Fellbaum(2010)]{poli_wordnet_2010}
Christiane Fellbaum.
\newblock {WordNet}.
\newblock In Roberto Poli, Michael Healy, and Achilles Kameas, editors, \emph{Theory and {Applications} of {Ontology}: {Computer} {Applications}}, pages 231--243. Springer Netherlands, Dordrecht, 2010.
\newblock ISBN 978-90-481-8846-8 978-90-481-8847-5.
\newblock \doi{10.1007/978-90-481-8847-5_10}.
\newblock URL \url{https://link.springer.com/10.1007/978-90-481-8847-5_10}.

\bibitem[Feng and Steinhardt(2023)]{feng2023language}
Jiahai Feng and Jacob Steinhardt.
\newblock How do language models bind entities in context?
\newblock \emph{arXiv preprint arXiv:2310.17191}, 2023.

\bibitem[Feng et~al.(2024)Feng, Russell, and Steinhardt]{feng2024monitoring}
Jiahai Feng, Stuart Russell, and Jacob Steinhardt.
\newblock Monitoring latent world states in language models with propositional probes.
\newblock \emph{arXiv preprint arXiv:2406.19501}, 2024.

\bibitem[Fodor(1975)]{fodor_language_1975}
Jerry~A. Fodor.
\newblock \emph{The language of thought}, volume~5.
\newblock Harvard university press Cambridge, MA, 1975.
\newblock URL \url{https://direct.mit.edu/books/edited-volume/3637/chapter-abstract/121528}.

\bibitem[Fodor and Pylyshyn(1988)]{fodor_connectionism_1988}
Jerry~A. Fodor and Zenon~W. Pylyshyn.
\newblock Connectionism and cognitive architecture: {A} critical analysis.
\newblock \emph{Cognition}, 28\penalty0 (1):\penalty0 3--71, mar 1988.
\newblock ISSN 0010-0277.
\newblock \doi{10.1016/0010-0277(88)90031-5}.
\newblock URL \url{https://www.sciencedirect.com/science/article/pii/0010027788900315}.

\bibitem[Francis and Kucera(1979)]{francis_brown_1979}
Winthrop~Nelson Francis and Henry Kucera.
\newblock \emph{Brown corpus maunal [ie manual]: {Manual} of information to accompany a standard corpus of present-day edited {American} {English} for use with digital computers}.
\newblock Department of Linguistics, Borwn University, 1979.

\bibitem[Friedman et~al.(2023)Friedman, Lampinen, Dixon, Chen, and Ghandeharioun]{friedman2023interpretability}
Dan Friedman, Andrew Lampinen, Lucas Dixon, Danqi Chen, and Asma Ghandeharioun.
\newblock Interpretability illusions in the generalization of simplified models.
\newblock \emph{arXiv preprint arXiv:2312.03656}, 2023.

\bibitem[Fukushima(1969)]{fukushima1969visual}
Kunihiko Fukushima.
\newblock Visual feature extraction by a multilayered network of analog threshold elements.
\newblock \emph{IEEE Transactions on Systems Science and Cybernetics}, 5\penalty0 (4):\penalty0 322--333, 1969.

\bibitem[Fyshe et~al.(2014)Fyshe, Talukdar, Murphy, and Mitchell]{fyshe_interpretable_2014}
Alona Fyshe, Partha~P. Talukdar, Brian Murphy, and Tom~M. Mitchell.
\newblock Interpretable {Semantic} {Vectors} from a {Joint} {Model} of {Brain}- and {Text}- {Based} {Meaning}.
\newblock In Kristina Toutanova and Hua Wu, editors, \emph{Proceedings of the 52nd {Annual} {Meeting} of the {Association} for {Computational} {Linguistics} ({Volume} 1: {Long} {Papers})}, pages 489--499, Baltimore, Maryland, jun 2014. Association for Computational Linguistics.
\newblock \doi{10.3115/v1/P14-1046}.
\newblock URL \url{https://aclanthology.org/P14-1046}.

\bibitem[Fyshe et~al.(2015)Fyshe, Wehbe, Talukdar, Murphy, and Mitchell]{fyshe_compositional_2015}
Alona Fyshe, Leila Wehbe, Partha~P. Talukdar, Brian Murphy, and Tom~M. Mitchell.
\newblock A {Compositional} and {Interpretable} {Semantic} {Space}.
\newblock In Rada Mihalcea, Joyce Chai, and Anoop Sarkar, editors, \emph{Proceedings of the 2015 {Conference} of the {North} {American} {Chapter} of the {Association} for {Computational} {Linguistics}: {Human} {Language} {Technologies}}, pages 32--41, Denver, Colorado, may 2015. Association for Computational Linguistics.
\newblock \doi{10.3115/v1/N15-1004}.
\newblock URL \url{https://aclanthology.org/N15-1004}.

\bibitem[Ganguli and Sompolinsky(2012)]{ganguli_compressed_2012}
Surya Ganguli and Haim Sompolinsky.
\newblock Compressed {Sensing}, {Sparsity}, and {Dimensionality} in {Neuronal} {Information} {Processing} and {Data} {Analysis}.
\newblock \emph{Annual Review of Neuroscience}, 35\penalty0 (Volume 35, 2012):\penalty0 485--508, jul 2012.
\newblock ISSN 0147-006X, 1545-4126.
\newblock \doi{10.1146/annurev-neuro-062111-150410}.
\newblock URL \url{https://www.annualreviews.org/content/journals/10.1146/annurev-neuro-062111-150410}.
\newblock Publisher: Annual Reviews.

\bibitem[Gao et~al.(2024)Gao, Goh, and Sutskever]{gao_scaling_2024}
Leo Gao, Gabriel Goh, and Ilya Sutskever.
\newblock Scaling and evaluating sparse autoencoders, jun 2024.
\newblock URL \url{https://cdn.openai.com/papers/sparse-autoencoders.pdf}.

\bibitem[Geirhos et~al.(2020)Geirhos, Jacobsen, Michaelis, Zemel, Brendel, Bethge, and Wichmann]{geirhos2020shortcut}
Robert Geirhos, J{\"o}rn-Henrik Jacobsen, Claudio Michaelis, Richard Zemel, Wieland Brendel, Matthias Bethge, and Felix~A Wichmann.
\newblock Shortcut learning in deep neural networks.
\newblock \emph{Nature Machine Intelligence}, 2\penalty0 (11):\penalty0 665--673, 2020.

\bibitem[Ghorbani et~al.(2019)Ghorbani, Wexler, Zou, and Kim]{ghorbani2019towards}
Amirata Ghorbani, James Wexler, James~Y Zou, and Been Kim.
\newblock Towards automatic concept-based explanations.
\newblock \emph{Advances in neural information processing systems}, 32, 2019.

\bibitem[Gilbert et~al.(2002)Gilbert, Guha, Indyk, Muthukrishnan, and Strauss]{gilbert2002near}
Anna~C Gilbert, Sudipto Guha, Piotr Indyk, Shanmugavelayutham Muthukrishnan, and Martin Strauss.
\newblock Near-optimal sparse fourier representations via sampling.
\newblock In \emph{Proceedings of the thiry-fourth annual ACM symposium on Theory of computing}, pages 152--161, 2002.

\bibitem[Gilpin et~al.(2018)Gilpin, Bau, Yuan, Bajwa, Specter, and Kagal]{gilpin2018explaining}
Leilani~H Gilpin, David Bau, Ben~Z Yuan, Ayesha Bajwa, Michael Specter, and Lalana Kagal.
\newblock Explaining explanations: An overview of interpretability of machine learning.
\newblock In \emph{2018 IEEE 5th International Conference on data science and advanced analytics (DSAA)}, pages 80--89. IEEE, 2018.

\bibitem[Gleichman and Eldar(2011)]{gleichman2011blind}
Sivan Gleichman and Yonina~C Eldar.
\newblock Blind compressed sensing.
\newblock \emph{IEEE Transactions on Information Theory}, 57\penalty0 (10):\penalty0 6958--6975, 2011.

\bibitem[Gross et~al.(1972)Gross, Rocha-Miranda, and Bender]{gross1972visual}
Charles~G Gross, CE~de Rocha-Miranda, and DB~Bender.
\newblock Visual properties of neurons in inferotemporal cortex of the macaque.
\newblock \emph{Journal of neurophysiology}, 35\penalty0 (1):\penalty0 96--111, 1972.

\bibitem[Higgins et~al.(2018)Higgins, Amos, Pfau, Racaniere, Matthey, Rezende, and Lerchner]{higgins_towards_2018}
Irina Higgins, David Amos, David Pfau, Sebastien Racaniere, Loic Matthey, Danilo Rezende, and Alexander Lerchner.
\newblock Towards a {Definition} of {Disentangled} {Representations}.
\newblock \emph{arXiv:1812.02230 [cs, stat]}, dec 2018.
\newblock URL \url{http://arxiv.org/abs/1812.02230}.
\newblock arXiv: 1812.02230.

\bibitem[Hillar and Sommer(2015{\natexlab{a}})]{hillar2015short}
Christopher~J. Hillar and Friedrich~T. Sommer.
\newblock When {Can} {Dictionary} {Learning} {Uniquely} {Recover} {Sparse} {Data} {From} {Subsamples}?
\newblock \emph{IEEE Transactions on Information Theory}, 61\penalty0 (11):\penalty0 6290--6297, nov 2015{\natexlab{a}}.
\newblock ISSN 1557-9654.
\newblock \doi{10.1109/TIT.2015.2460238}.
\newblock Conference Name: IEEE Transactions on Information Theory.

\bibitem[Hillar and Sommer(2015{\natexlab{b}})]{hillar_when_2015}
Christopher~J. Hillar and Friedrich~T. Sommer.
\newblock When {Can} {Dictionary} {Learning} {Uniquely} {Recover} {Sparse} {Data} {From} {Subsamples}?
\newblock \emph{IEEE Transactions on Information Theory}, 61\penalty0 (11):\penalty0 6290--6297, nov 2015{\natexlab{b}}.
\newblock ISSN 1557-9654.
\newblock \doi{10.1109/TIT.2015.2460238}.
\newblock URL \url{https://ieeexplore.ieee.org/abstract/document/7165675?casa_token=75WtBxUgMQgAAAAA:eek6hgNU40XdBfDgDAT29nt-lxXVndIdsbl9XZZLRnKIT8pnv8UHaxNhOKZQUV9pItE5IfmqlQ}.
\newblock Conference Name: IEEE Transactions on Information Theory.

\bibitem[Honkela et~al.(2003)Honkela, Hyvärinen, and Väyrynen]{honkela_emergence_2003}
Timo Honkela, Aapo Hyvärinen, and Jaakko Väyrynen.
\newblock \emph{Emergence of linguistic representations by independent component analysis}.
\newblock Helsinki University of Technology, 2003.
\newblock URL \url{http://research.ics.aalto.fi/publications/jjvayryn/Honkela03tr.pdf}.

\bibitem[Honkela et~al.(2010)Honkela, Hyvärinen, and Väyrynen]{honkela_wordicaemergence_2010}
Timo Honkela, Aapo Hyvärinen, and Jaakko~J. Väyrynen.
\newblock {WordICA}—emergence of linguistic representations for words by independent component analysis.
\newblock \emph{Natural Language Engineering}, 16\penalty0 (3):\penalty0 277--308, 2010.
\newblock Publisher: Cambridge University Press.

\bibitem[Hubel and Wiesel(1962)]{hubel1962receptive}
David~H Hubel and Torsten~N Wiesel.
\newblock Receptive fields, binocular interaction and functional architecture in the cat's visual cortex.
\newblock \emph{The Journal of physiology}, 160\penalty0 (1):\penalty0 106, 1962.

\bibitem[Huh et~al.(2024)Huh, Cheung, Wang, and Isola]{huh2024platonic}
Minyoung Huh, Brian Cheung, Tongzhou Wang, and Phillip Isola.
\newblock The platonic representation hypothesis.
\newblock \emph{arXiv preprint arXiv:2405.07987}, 2024.

\bibitem[Hyttinen et~al.(2022)Hyttinen, Pacela, and Hyv{\"a}rinen]{hyttinen2022binary}
Antti Hyttinen, Vit{\'o}ria~Barin Pacela, and Aapo Hyv{\"a}rinen.
\newblock Binary independent component analysis: a non-stationarity-based approach.
\newblock In \emph{Uncertainty in Artificial Intelligence}, pages 874--884. PMLR, 2022.

\bibitem[Hyv\"arinen and Morioka(2016)]{Hyva16NIPS}
A.~Hyv\"arinen and H.~Morioka.
\newblock Unsupervised feature extraction by time-contrastive learning and nonlinear {ICA}.
\newblock In \emph{Advances in Neural Information Processing Systems (NIPS2016)}, Barcelona, Spain, 2016.

\bibitem[Hyvarinen and Morioka(2017)]{hyvarinen_nonlinear_2017}
Aapo Hyvarinen and Hiroshi Morioka.
\newblock Nonlinear {ICA} of {Temporally} {Dependent} {Stationary} {Sources}.
\newblock In \emph{Artificial {Intelligence} and {Statistics}}, pages 460--469. PMLR, apr 2017.
\newblock URL \url{http://proceedings.mlr.press/v54/hyvarinen17a.html}.
\newblock ISSN: 2640-3498.

\bibitem[Hyv{\"a}rinen and Pajunen(1999)]{hyvarinen1999nonlinear}
Aapo Hyv{\"a}rinen and Petteri Pajunen.
\newblock Nonlinear independent component analysis: Existence and uniqueness results.
\newblock \emph{Neural networks}, 12\penalty0 (3):\penalty0 429--439, 1999.

\bibitem[Hyvarinen et~al.(2019{\natexlab{a}})Hyvarinen, Sasaki, and Turner]{hyvarinen2019nonlinear}
Aapo Hyvarinen, Hiroaki Sasaki, and Richard Turner.
\newblock Nonlinear ica using auxiliary variables and generalized contrastive learning.
\newblock In \emph{The 22nd International Conference on Artificial Intelligence and Statistics}, pages 859--868. PMLR, 2019{\natexlab{a}}.

\bibitem[Hyvarinen et~al.(2019{\natexlab{b}})Hyvarinen, Sasaki, and Turner]{hyvarinen_nonlinear_2019}
Aapo Hyvarinen, Hiroaki Sasaki, and Richard Turner.
\newblock Nonlinear {ICA} {Using} {Auxiliary} {Variables} and {Generalized} {Contrastive} {Learning}.
\newblock In \emph{Proceedings of the {Twenty}-{Second} {International} {Conference} on {Artificial} {Intelligence} and {Statistics}}, pages 859--868. PMLR, apr 2019{\natexlab{b}}.
\newblock URL \url{https://proceedings.mlr.press/v89/hyvarinen19a.html}.
\newblock arXiv: 1805.08651.

\bibitem[Hyvärinen and Oja(2000)]{hyvarinen_independent_2000}
A.~Hyvärinen and E.~Oja.
\newblock Independent component analysis: algorithms and applications.
\newblock \emph{Neural Networks}, 13\penalty0 (4):\penalty0 411--430, jun 2000.
\newblock ISSN 0893-6080.
\newblock \doi{10.1016/S0893-6080(00)00026-5}.
\newblock URL \url{https://www.sciencedirect.com/science/article/pii/S0893608000000265}.

\bibitem[Hyvärinen et~al.(2023)Hyvärinen, Khemakhem, and Monti]{hyvarinen_identifiability_2023}
Aapo Hyvärinen, Ilyes Khemakhem, and Ricardo Monti.
\newblock Identifiability of latent-variable and structural-equation models: from linear to nonlinear, may 2023.
\newblock URL \url{http://arxiv.org/abs/2302.02672}.
\newblock arXiv:2302.02672 [cs, stat].

\bibitem[Illingworth(1991)]{ill1991}
Valerie Illingworth.
\newblock The {Penguin} dictionary of physics.
\newblock \emph{Penguin Books 2nd edition}, 1991.
\newblock URL \url{https://cir.nii.ac.jp/crid/1130000797436622208}.

\bibitem[Jang and Myaeng(2017)]{jang_elucidating_2017}
Kyoung-Rok Jang and Sung-Hyon Myaeng.
\newblock Elucidating {Conceptual} {Properties} from {Word} {Embeddings}.
\newblock In Jose Camacho-Collados and Mohammad~Taher Pilehvar, editors, \emph{Proceedings of the 1st {Workshop} on {Sense}, {Concept} and {Entity} {Representations} and their {Applications}}, pages 91--95, Valencia, Spain, apr 2017. Association for Computational Linguistics.
\newblock \doi{10.18653/v1/W17-1911}.
\newblock URL \url{https://aclanthology.org/W17-1911}.

\bibitem[Juang et~al.(2024)Juang, Paulo, Drori, and Belrose]{eleutherOpenSource}
Caden Juang, Gonçalo Paulo, Jacob Drori, and Nora Belrose.
\newblock Open source automated interpretability for sparse autoencoder features, 2024.
\newblock URL \url{https://blog.eleuther.ai/autointerp/}.
\newblock [Accessed 29-09-2024].

\bibitem[Keurti et~al.(2023)Keurti, Reizinger, Schölkopf, and Brendel]{keurti_desiderata_2023}
Hamza Keurti, Patrik Reizinger, Bernhard Schölkopf, and Wieland Brendel.
\newblock Desiderata for {Representation} {Learning} from {Identifiability}, {Disentanglement}, and {Group}-{Structuredness}, jun 2023.
\newblock URL \url{https://openreview.net/forum?id=r6C86JjuiW}.

\bibitem[Khemakhem et~al.(2020{\natexlab{a}})Khemakhem, Kingma, Monti, and Hyvarinen]{khemakhem_variational_2020}
Ilyes Khemakhem, Diederik Kingma, Ricardo Monti, and Aapo Hyvarinen.
\newblock Variational {Autoencoders} and {Nonlinear} {ICA}: {A} {Unifying} {Framework}.
\newblock In \emph{International {Conference} on {Artificial} {Intelligence} and {Statistics}}, pages 2207--2217. PMLR, jun 2020{\natexlab{a}}.
\newblock URL \url{http://proceedings.mlr.press/v108/khemakhem20a.html}.
\newblock ISSN: 2640-3498.

\bibitem[Khemakhem et~al.(2020{\natexlab{b}})Khemakhem, Monti, Kingma, and Hyvärinen]{khemakhem_ice-beem_2020}
Ilyes Khemakhem, Ricardo~Pio Monti, Diederik~P. Kingma, and Aapo Hyvärinen.
\newblock {ICE}-{BeeM}: {Identifiable} {Conditional} {Energy}-{Based} {Deep} {Models} {Based} on {Nonlinear} {ICA}.
\newblock \emph{arXiv:2002.11537 [cs, stat]}, oct 2020{\natexlab{b}}.
\newblock URL \url{http://arxiv.org/abs/2002.11537}.
\newblock arXiv: 2002.11537.

\bibitem[Khosla et~al.(2024)Khosla, Williams, McDermott, and Kanwisher]{khosla_privileged_2024}
Meenakshi Khosla, Alex~H. Williams, Josh McDermott, and Nancy Kanwisher.
\newblock Privileged representational axes in biological and artificial neural networks, jun 2024.
\newblock URL \url{https://www.biorxiv.org/content/10.1101/2024.06.20.599957v1}.
\newblock Pages: 2024.06.20.599957 Section: New Results.

\bibitem[Kingma(2013)]{kingma2013auto}
Diederik~P Kingma.
\newblock Auto-encoding variational bayes.
\newblock \emph{arXiv preprint arXiv:1312.6114}, 2013.

\bibitem[Kleene(1951)]{kleene1951representationof}
SC~Kleene.
\newblock Representationof events in nerve nets and finite automata.
\newblock \emph{CE Shannon and J. McCarthy}, 1951.

\bibitem[Klindt et~al.(2020)Klindt, Schott, Sharma, Ustyuzhaninov, Brendel, Bethge, and Paiton]{klindt2020towards}
David Klindt, Lukas Schott, Yash Sharma, Ivan Ustyuzhaninov, Wieland Brendel, Matthias Bethge, and Dylan Paiton.
\newblock Towards nonlinear disentanglement in natural data with temporal sparse coding.
\newblock \emph{arXiv preprint arXiv:2007.10930}, 2020.

\bibitem[Klindt et~al.(2021)Klindt, Schott, Sharma, Ustyuzhaninov, Brendel, Bethge, and Paiton]{klindt2021towards}
David Klindt, Lukas Schott, Yash Sharma, Ivan Ustyuzhaninov, Wieland Brendel, Matthias Bethge, and Dylan Paiton.
\newblock Towards nonlinear disentanglement in natural data with temporal sparse coding.
\newblock \emph{ICLR}, 2021.

\bibitem[Klindt et~al.(2023)Klindt, Sanborn, Acosta, Poitevin, and Miolane]{klindt_identifying_2023}
David Klindt, Sophia Sanborn, Francisco Acosta, Frédéric Poitevin, and Nina Miolane.
\newblock Identifying {Interpretable} {Visual} {Features} in {Artificial} and {Biological} {Neural} {Systems}, oct 2023.
\newblock URL \url{http://arxiv.org/abs/2310.11431}.
\newblock arXiv:2310.11431 [cs, stat].

\bibitem[Klindt et~al.(2024)Klindt, Hyv{\"a}rinen, Levy, Miolane, and Poitevin]{klindt2024towards}
David~A Klindt, Aapo Hyv{\"a}rinen, Axel Levy, Nina Miolane, and Fr{\'e}d{\'e}ric Poitevin.
\newblock Towards interpretable cryo-em: disentangling latent spaces of molecular conformations.
\newblock \emph{Frontiers in Molecular Biosciences}, 11:\penalty0 1393564, 2024.

\bibitem[Kohonen(1982)]{kohonen1982self}
Teuvo Kohonen.
\newblock Self-organized formation of topologically correct feature maps.
\newblock \emph{Biological cybernetics}, 43\penalty0 (1):\penalty0 59--69, 1982.

\bibitem[Lachapelle et~al.(2023)Lachapelle, Deleu, Mahajan, Mitliagkas, Bengio, Lacoste-Julien, and Bertrand]{lachapelle_synergies_2023}
Sebastien Lachapelle, Tristan Deleu, Divyat Mahajan, Ioannis Mitliagkas, Yoshua Bengio, Simon Lacoste-Julien, and Quentin Bertrand.
\newblock Synergies between {Disentanglement} and {Sparsity}: {Generalization} and {Identifiability} in {Multi}-{Task} {Learning}.
\newblock In \emph{Proceedings of the 40th {International} {Conference} on {Machine} {Learning}}, pages 18171--18206. PMLR, jul 2023.
\newblock URL \url{https://proceedings.mlr.press/v202/lachapelle23a.html}.
\newblock ISSN: 2640-3498.

\bibitem[Lake et~al.(2017)Lake, Ullman, Tenenbaum, and Gershman]{lake2017building}
Brenden~M Lake, Tomer~D Ullman, Joshua~B Tenenbaum, and Samuel~J Gershman.
\newblock Building machines that learn and think like people.
\newblock \emph{Behavioral and brain sciences}, 40:\penalty0 e253, 2017.

\bibitem[Lan et~al.(2024)Lan, Torr, Meek, Khakzar, Krueger, and Barez]{lan_sparse_2024}
Michael Lan, Philip Torr, Austin Meek, Ashkan Khakzar, David Krueger, and Fazl Barez.
\newblock Sparse {Autoencoders} {Reveal} {Universal} {Feature} {Spaces} {Across} {Large} {Language} {Models}, oct 2024.
\newblock URL \url{http://arxiv.org/abs/2410.06981}.
\newblock arXiv:2410.06981.

\bibitem[Leavitt and Morcos(2020)]{leavitt_towards_2020}
Matthew~L. Leavitt and Ari Morcos.
\newblock Towards falsifiable interpretability research, oct 2020.
\newblock URL \url{http://arxiv.org/abs/2010.12016}.
\newblock arXiv:2010.12016 [cs, stat].

\bibitem[Li et~al.(2019)Li, Brendel, Walker, Cobos, Muhammad, Reimer, Bethge, Sinz, Pitkow, and Tolias]{li_learning_2019}
Zhe Li, Wieland Brendel, Edgar Walker, Erick Cobos, Taliah Muhammad, Jacob Reimer, Matthias Bethge, Fabian Sinz, Zachary Pitkow, and Andreas Tolias.
\newblock Learning from brains how to regularize machines.
\newblock In \emph{Advances in {Neural} {Information} {Processing} {Systems}}, volume~32, pages 4245--4252. Curran Associates, Inc., jul 2019.
\newblock \doi{10.1609/aaai.v33i01.33014245}.
\newblock URL \url{https://proceedings.neurips.cc/paper/2019/hash/70117ee3c0b15a2950f1e82a215e812b-Abstract.html}.
\newblock Number: 01.

\bibitem[Lim et~al.(2024)Lim, Choi, Choo, and Schneider]{lim2024sparseautoencodersrevealselective}
Hyesu Lim, Jinho Choi, Jaegul Choo, and Steffen Schneider.
\newblock Sparse autoencoders reveal selective remapping of visual concepts during adaptation, 2024.
\newblock URL \url{https://arxiv.org/abs/2412.05276}.

\bibitem[Linardatos et~al.(2020)Linardatos, Papastefanopoulos, and Kotsiantis]{linardatos2020explainable}
Pantelis Linardatos, Vasilis Papastefanopoulos, and Sotiris Kotsiantis.
\newblock Explainable ai: A review of machine learning interpretability methods.
\newblock \emph{Entropy}, 23\penalty0 (1):\penalty0 18, 2020.

\bibitem[Lindsey et~al.(2024)Lindsey, Templeton, Marcus, Conerly, Batson, and Olah]{lindsey2024crosscoder}
Jack Lindsey, Adly Templeton, Jonathan Marcus, Thomas Conerly, Joshua Batson, and Christopher Olah.
\newblock \emph{Sparse Crosscoders for Cross-Layer Features and Model Diffing}, 2024.
\newblock URL \url{https://transformer-circuits.pub/2024/crosscoders/index.html}.

\bibitem[Locatello et~al.(2019{\natexlab{a}})Locatello, Bauer, Lucic, Raetsch, Gelly, Sch{\"o}lkopf, and Bachem]{locatello2019challenging}
Francesco Locatello, Stefan Bauer, Mario Lucic, Gunnar Raetsch, Sylvain Gelly, Bernhard Sch{\"o}lkopf, and Olivier Bachem.
\newblock Challenging common assumptions in the unsupervised learning of disentangled representations.
\newblock In \emph{international conference on machine learning}, pages 4114--4124. PMLR, 2019{\natexlab{a}}.

\bibitem[Locatello et~al.(2019{\natexlab{b}})Locatello, Bauer, Lucic, Raetsch, Gelly, Schölkopf, and Bachem]{locatello_challenging_2019}
Francesco Locatello, Stefan Bauer, Mario Lucic, Gunnar Raetsch, Sylvain Gelly, Bernhard Schölkopf, and Olivier Bachem.
\newblock Challenging {Common} {Assumptions} in the {Unsupervised} {Learning} of {Disentangled} {Representations}.
\newblock In \emph{International {Conference} on {Machine} {Learning}}, pages 4114--4124. PMLR, may 2019{\natexlab{b}}.
\newblock URL \url{http://proceedings.mlr.press/v97/locatello19a.html}.
\newblock ISSN: 2640-3498.

\bibitem[Mazor and Laurent(2005)]{mazor2005transient}
Ofer Mazor and Gilles Laurent.
\newblock Transient dynamics versus fixed points in odor representations by locust antennal lobe projection neurons.
\newblock \emph{Neuron}, 48\penalty0 (4):\penalty0 661--673, 2005.

\bibitem[McClure and Kriegeskorte(2016)]{mcclure_representational_2016}
Patrick McClure and Nikolaus Kriegeskorte.
\newblock Representational {Distance} {Learning} for {Deep} {Neural} {Networks}.
\newblock \emph{Frontiers in Computational Neuroscience}, 10, dec 2016.
\newblock ISSN 1662-5188.
\newblock \doi{10.3389/fncom.2016.00131}.
\newblock URL \url{https://www.frontiersin.org/articles/10.3389/fncom.2016.00131}.
\newblock Publisher: Frontiers.

\bibitem[McCulloch and Pitts(1943)]{mcculloch1943logical}
Warren~S McCulloch and Walter Pitts.
\newblock A logical calculus of the ideas immanent in nervous activity.
\newblock \emph{The bulletin of mathematical biophysics}, 5:\penalty0 115--133, 1943.

\bibitem[Mead(1992)]{mead1992review}
Al~Mead.
\newblock Review of the development of multidimensional scaling methods.
\newblock \emph{Journal of the Royal Statistical Society: Series D (The Statistician)}, 41\penalty0 (1):\penalty0 27--39, 1992.

\bibitem[Mikolov(2013)]{mikolov2013efficient}
Tomas Mikolov.
\newblock Efficient estimation of word representations in vector space.
\newblock \emph{arXiv preprint arXiv:1301.3781}, 2013.

\bibitem[Mikolov et~al.(2013)Mikolov, Chen, Corrado, and Dean]{mikolov_efficient_2013}
Tomas Mikolov, Kai Chen, Greg Corrado, and Jeffrey Dean.
\newblock Efficient {Estimation} of {Word} {Representations} in {Vector} {Space}, sep 2013.
\newblock URL \url{http://arxiv.org/abs/1301.3781}.
\newblock arXiv:1301.3781 [cs].

\bibitem[Morioka and Hyvarinen(2023)]{morioka_connectivity-contrastive_2023}
Hiroshi Morioka and Aapo Hyvarinen.
\newblock Connectivity-contrastive learning: {Combining} causal discovery and representation learning for multimodal data.
\newblock In \emph{Proceedings of {The} 26th {International} {Conference} on {Artificial} {Intelligence} and {Statistics}}, pages 3399--3426. PMLR, apr 2023.
\newblock URL \url{https://proceedings.mlr.press/v206/morioka23a.html}.
\newblock ISSN: 2640-3498.

\bibitem[Morioka et~al.(2021)Morioka, Hälvä, and Hyvärinen]{morioka_independent_2021}
Hiroshi Morioka, Hermanni Hälvä, and Aapo Hyvärinen.
\newblock Independent {Innovation} {Analysis} for {Nonlinear} {Vector} {Autoregressive} {Process}.
\newblock \emph{arXiv:2006.10944 [cs, stat]}, feb 2021.
\newblock URL \url{https://arxiv.org/abs/2006.10944}.
\newblock arXiv: 2006.10944.

\bibitem[Murphy et~al.(2012)Murphy, Talukdar, and Mitchell]{murphy_learning_2012}
Brian Murphy, Partha Talukdar, and Tom Mitchell.
\newblock Learning {Effective} and {Interpretable} {Semantic} {Models} using {Non}-{Negative} {Sparse} {Embedding}.
\newblock \emph{Proceedings of COLING 2012}, Proceedings of COLING 2012:\penalty0 1933--1950, 2012.
\newblock URL \url{https://aclanthology.org/C12-1118.pdf}.

\bibitem[Natarajan(1995)]{natarajan1995sparse}
Balas~Kausik Natarajan.
\newblock Sparse approximate solutions to linear systems.
\newblock \emph{SIAM journal on computing}, 24\penalty0 (2):\penalty0 227--234, 1995.

\bibitem[Ng et~al.(2011)]{ng2011sparse}
Andrew Ng et~al.
\newblock Sparse autoencoder.
\newblock \emph{CS294A Lecture notes}, 72\penalty0 (2011):\penalty0 1--19, 2011.

\bibitem[Olah et~al.(2017)Olah, Mordvintsev, and Schubert]{olah_feature_2017}
Chris Olah, Alexander Mordvintsev, and Ludwig Schubert.
\newblock Feature visualization.
\newblock \emph{Distill}, 2\penalty0 (11):\penalty0 e7, 2017.
\newblock URL \url{https://distill.pub/2017/feature-visualization/}.

\bibitem[Olshausen(2008)]{olshausen08sc_ica}
Bruno Olshausen.
\newblock Sparse coding and `ica'.
\newblock \emph{online}, 2008.
\newblock URL \url{https://redwood.berkeley.edu/wp-content/uploads/2018/08/sparse-coding-ICA.pdf}.

\bibitem[Olshausen and Field(1996)]{olshausen_emergence_1996}
Bruno~A. Olshausen and David~J. Field.
\newblock Emergence of simple-cell receptive field properties by learning a sparse code for natural images.
\newblock \emph{Nature}, 381\penalty0 (6583):\penalty0 607--609, jun 1996.
\newblock ISSN 1476-4687.
\newblock \doi{10.1038/381607a0}.
\newblock URL \url{https://www.nature.com/articles/381607a0}.
\newblock Publisher: Nature Publishing Group.

\bibitem[Olsson et~al.(2022)Olsson, Elhage, Nanda, Joseph, DasSarma, Henighan, Mann, Askell, Bai, Chen, Conerly, Drain, Ganguli, Hatfield-Dodds, Hernandez, Johnston, Jones, Kernion, Lovitt, Ndousse, Amodei, Brown, Clark, Kaplan, McCandlish, and Olah]{olsson_-context_2022}
Catherine Olsson, Nelson Elhage, Neel Nanda, Nicholas Joseph, Nova DasSarma, Tom Henighan, Ben Mann, Amanda Askell, Yuntao Bai, Anna Chen, Tom Conerly, Dawn Drain, Deep Ganguli, Zac Hatfield-Dodds, Danny Hernandez, Scott Johnston, Andy Jones, Jackson Kernion, Liane Lovitt, Kamal Ndousse, Dario Amodei, Tom Brown, Jack Clark, Jared Kaplan, Sam McCandlish, and Chris Olah.
\newblock In-context {Learning} and {Induction} {Heads}, sep 2022.
\newblock URL \url{http://arxiv.org/abs/2209.11895}.
\newblock arXiv:2209.11895 [cs].

\bibitem[O'Neill and Bui(2024)]{oneill_sparse_2024}
Charles O'Neill and Thang Bui.
\newblock Sparse {Autoencoders} {Enable} {Scalable} and {Reliable} {Circuit} {Identification} in {Language} {Models}, may 2024.
\newblock URL \url{http://arxiv.org/abs/2405.12522}.
\newblock arXiv:2405.12522 [cs].

\bibitem[O'Neill et~al.(2025)O'Neill, Gumran, and Klindt]{oneill2025computeoptimalinferenceprovable}
Charles O'Neill, Alim Gumran, and David Klindt.
\newblock Compute optimal inference and provable amortisation gap in sparse autoencoders, 2025.
\newblock URL \url{https://arxiv.org/abs/2411.13117}.

\bibitem[Paiton et~al.(2020)Paiton, Frye, Lundquist, Bowen, Zarcone, and Olshausen]{paiton2020selectivity}
Dylan~M Paiton, Charles~G Frye, Sheng~Y Lundquist, Joel~D Bowen, Ryan Zarcone, and Bruno~A Olshausen.
\newblock Selectivity and robustness of sparse coding networks.
\newblock \emph{Journal of vision}, 20\penalty0 (12):\penalty0 10--10, 2020.

\bibitem[Park et~al.(2024{\natexlab{a}})Park, Choe, Jiang, and Veitch]{park2024geometry}
Kiho Park, Yo~Joong Choe, Yibo Jiang, and Victor Veitch.
\newblock The geometry of categorical and hierarchical concepts in large language models.
\newblock \emph{arXiv preprint arXiv:2406.01506}, 2024{\natexlab{a}}.

\bibitem[Park et~al.(2024{\natexlab{b}})Park, Choe, and Veitch]{park_linear_2024}
Kiho Park, Yo~Joong Choe, and Victor Veitch.
\newblock The {Linear} {Representation} {Hypothesis} and the {Geometry} of {Large} {Language} {Models}, jul 2024{\natexlab{b}}.
\newblock URL \url{http://arxiv.org/abs/2311.03658}.
\newblock arXiv:2311.03658 [cs, stat].

\bibitem[Paulo and Belrose(2025)]{paulo2025sparse}
Gon{\c{c}}alo Paulo and Nora Belrose.
\newblock Sparse autoencoders trained on the same data learn different features.
\newblock \emph{arXiv preprint arXiv:2501.16615}, 2025.

\bibitem[Pennington et~al.(2014)Pennington, Socher, and Manning]{pennington_glove_2014}
Jeffrey Pennington, Richard Socher, and Christopher~D. Manning.
\newblock Glove: {Global} vectors for word representation.
\newblock In \emph{Proceedings of the 2014 conference on empirical methods in natural language processing ({EMNLP})}, pages 1532--1543, 2014.
\newblock URL \url{https://aclanthology.org/D14-1162.pdf}.

\bibitem[Pfau et~al.(2020)Pfau, Higgins, Botev, and Racanière]{pfau_disentangling_2020}
David Pfau, Irina Higgins, Alex Botev, and Sébastien Racanière.
\newblock Disentangling by {Subspace} {Diffusion}.
\newblock In \emph{Advances in {Neural} {Information} {Processing} {Systems}}, volume~33, pages 17403--17415. Curran Associates, Inc., 2020.
\newblock URL \url{https://proceedings.neurips.cc/paper/2020/hash/c9f029a6a1b20a8408f372351b321dd8-Abstract.html}.

\bibitem[Quian et~al.(2005)Quian, Reddy, Kreiman, Koch, and Fried]{quian_invariant_2005}
Rodrigo Quian, Leila Reddy, Gabriel Kreiman, Christof Koch, and Itzhak Fried.
\newblock Invariant {Visual} {Representation} by {Single} {Neurons} in the {Human} {Brain}.
\newblock \emph{Nature}, 435:\penalty0 1102--7, jul 2005.
\newblock \doi{10.1038/nature03687}.

\bibitem[Radford et~al.(2021)Radford, Kim, Hallacy, Ramesh, Goh, Agarwal, Sastry, Askell, Mishkin, Clark, Krueger, and Sutskever]{radford_learning_2021}
Alec Radford, Jong~Wook Kim, Chris Hallacy, Aditya Ramesh, Gabriel Goh, Sandhini Agarwal, Girish Sastry, Amanda Askell, Pamela Mishkin, Jack Clark, Gretchen Krueger, and Ilya Sutskever.
\newblock Learning {Transferable} {Visual} {Models} {From} {Natural} {Language} {Supervision}, feb 2021.
\newblock URL \url{http://arxiv.org/abs/2103.00020}.
\newblock arXiv:2103.00020 [cs].

\bibitem[Rajamanoharan et~al.(2024{\natexlab{a}})Rajamanoharan, Conmy, Smith, Lieberum, Varma, Kramár, Shah, and Nanda]{rajamanoharan_improving_2024}
Senthooran Rajamanoharan, Arthur Conmy, Lewis Smith, Tom Lieberum, Vikrant Varma, János Kramár, Rohin Shah, and Neel Nanda.
\newblock Improving {Dictionary} {Learning} with {Gated} {Sparse} {Autoencoders}, apr 2024{\natexlab{a}}.
\newblock URL \url{http://arxiv.org/abs/2404.16014}.
\newblock arXiv:2404.16014 [cs].

\bibitem[Rajamanoharan et~al.(2024{\natexlab{b}})Rajamanoharan, Lieberum, Sonnerat, Conmy, Varma, Kramár, and Nanda]{rajamanoharan2024jumpingaheadimprovingreconstruction}
Senthooran Rajamanoharan, Tom Lieberum, Nicolas Sonnerat, Arthur Conmy, Vikrant Varma, János Kramár, and Neel Nanda.
\newblock Jumping ahead: Improving reconstruction fidelity with jumprelu sparse autoencoders, 2024{\natexlab{b}}.
\newblock URL \url{https://arxiv.org/abs/2407.14435}.

\bibitem[Reizinger et~al.(2024)Reizinger, Bizeul, Juhos, Vogt, Balestriero, Brendel, and Klindt]{reizinger2024cross}
Patrik Reizinger, Alice Bizeul, Attila Juhos, Julia~E Vogt, Randall Balestriero, Wieland Brendel, and David Klindt.
\newblock Cross-entropy is all you need to invert the data generating process.
\newblock \emph{arXiv preprint arXiv:2410.21869}, 2024.

\bibitem[Rezende et~al.(2014)Rezende, Mohamed, and Wierstra]{rezende2014stochastic}
Danilo~Jimenez Rezende, Shakir Mohamed, and Daan Wierstra.
\newblock Stochastic backpropagation and approximate inference in deep generative models.
\newblock In \emph{International conference on machine learning}, pages 1278--1286. PMLR, 2014.

\bibitem[Rigotti et~al.(2013)Rigotti, Barak, Warden, Wang, Daw, Miller, and Fusi]{rigotti_importance_2013}
Mattia Rigotti, Omri Barak, Melissa~R. Warden, Xiao-Jing Wang, Nathaniel~D. Daw, Earl~K. Miller, and Stefano Fusi.
\newblock The importance of mixed selectivity in complex cognitive tasks.
\newblock \emph{Nature}, 497\penalty0 (7451):\penalty0 585--590, 2013.
\newblock URL \url{https://www.nature.com/articles/nature12160}.
\newblock Publisher: Nature Publishing Group UK London.

\bibitem[Ritter and Kohonen(1989)]{ritter_self-organizing_1989}
H.~Ritter and T.~Kohonen.
\newblock Self-organizing semantic maps.
\newblock \emph{Biological Cybernetics}, 61\penalty0 (4):\penalty0 241--254, aug 1989.
\newblock ISSN 1432-0770.
\newblock \doi{10.1007/BF00203171}.
\newblock URL \url{https://doi.org/10.1007/BF00203171}.

\bibitem[Rosenblatt(1958)]{rosenblatt1958perceptron}
Frank Rosenblatt.
\newblock The perceptron: a probabilistic model for information storage and organization in the brain.
\newblock \emph{Psychological review}, 65\penalty0 (6):\penalty0 386, 1958.

\bibitem[Rumelhart et~al.(1986)Rumelhart, Hinton, and McClelland]{rumelhart_general_1986}
David~E. Rumelhart, Geoffrey~E. Hinton, and James~L. McClelland.
\newblock A general framework for parallel distributed processing.
\newblock \emph{Parallel distributed processing: Explorations in the microstructure of cognition}, 1\penalty0 (45-76):\penalty0 26, 1986.
\newblock Publisher: Cambridge, MA: MIT Press.

\bibitem[Rusak et~al.(2024)Rusak, Reizinger, Juhos, Bringmann, Zimmermann, and Brendel]{rusak_infonce_2024}
Evgenia Rusak, Patrik Reizinger, Attila Juhos, Oliver Bringmann, Roland~S. Zimmermann, and Wieland Brendel.
\newblock {InfoNCE}: {Identifying} the {Gap} {Between} {Theory} and {Practice}, jun 2024.
\newblock URL \url{http://arxiv.org/abs/2407.00143}.
\newblock arXiv:2407.00143 [cs, stat].

\bibitem[Saxena and Cunningham(2019)]{saxena2019towards}
Shreya Saxena and John~P Cunningham.
\newblock Towards the neural population doctrine.
\newblock \emph{Current opinion in neurobiology}, 55:\penalty0 103--111, 2019.

\bibitem[Schuster(2024)]{schuster2024can}
Viktoria Schuster.
\newblock Can sparse autoencoders make sense of latent representations?
\newblock \emph{arXiv preprint arXiv:2410.11468}, 2024.

\bibitem[Simon and Zou(2024)]{simon2024interplm}
Elana Simon and James Zou.
\newblock Interplm: Discovering interpretable features in protein language models via sparse autoencoders.
\newblock \emph{bioRxiv}, pages 2024--11, 2024.

\bibitem[Smith(2024)]{smith2024weak_lrh}
Lewis Smith.
\newblock The ‘strong’ feature hypothesis could be wrong, 2024.
\newblock URL \url{https://www.alignmentforum.org/posts/tojtPCCRpKLSHBdpn/ the-strong-feature-hypothesis-could-be-wrong}.
\newblock AI Alignment Forum.

\bibitem[Smolensky(1990)]{smolensky_tensor_1990}
Paul Smolensky.
\newblock Tensor product variable binding and the representation of symbolic structures in connectionist systems.
\newblock \emph{Artificial intelligence}, 46\penalty0 (1-2):\penalty0 159--216, 1990.
\newblock URL \url{https://www.sciencedirect.com/science/article/pii/000437029090007M}.
\newblock Publisher: Elsevier.

\bibitem[Snow et~al.(2006)Snow, Jurafsky, and Ng]{snow_semantic_2006}
Rion Snow, Dan Jurafsky, and Andrew~Y. Ng.
\newblock Semantic taxonomy induction from heterogenous evidence.
\newblock In \emph{Proceedings of the 21st international conference on computational linguistics and 44th annual meeting of the association for computational linguistics}, pages 801--808, 2006.
\newblock URL \url{https://aclanthology.org/P06-1101.pdf}.

\bibitem[Stojnic(2010)]{stojnic2010recovery}
Mihailo Stojnic.
\newblock Recovery thresholds for l1 optimization in binary compressed sensing.
\newblock In \emph{2010 IEEE International Symposium on Information Theory}, pages 1593--1597. IEEE, 2010.

\bibitem[Subramanian et~al.(2017)Subramanian, Pruthi, Jhamtani, Berg-Kirkpatrick, and Hovy]{subramanian_spine_2017}
Anant Subramanian, Danish Pruthi, Harsh Jhamtani, Taylor Berg-Kirkpatrick, and Eduard Hovy.
\newblock {SPINE}: {SParse} {Interpretable} {Neural} {Embeddings}, nov 2017.
\newblock URL \url{http://arxiv.org/abs/1711.08792}.
\newblock arXiv:1711.08792 [cs].

\bibitem[Templeton et~al.(2024{\natexlab{a}})Templeton, Conerly, Marcus, Lindsey, Bricken, and al.]{templeton_scaling_2024}
Adly Templeton, Tom Conerly, Jonathan Marcus, Jack Lindsey, Trenton Bricken, and et~al.
\newblock Scaling {Monosemanticity}: {Extracting} {Interpretable} {Features} from {Claude} 3 {Sonnet}, 2024{\natexlab{a}}.
\newblock URL \url{https://transformer-circuits.pub/2024/scaling-monosemanticity}.

\bibitem[Templeton et~al.(2024{\natexlab{b}})Templeton, Conerly, Marcus, Lindsey, Bricken, Chen, Pearce, Citro, Ameisen, Andy~Jones, Turner, McDougall, MacDiarmid, Tamkin, Durmus, Hume, Mosconi, Freeman, Sumers, Rees, Batson, Jermyn, Carter, Olah, and Henighan]{transformercircuitsScalingMonosemanticity}
Adly Templeton, Tom Conerly, Jonathan Marcus, Jack Lindsey, Trenton Bricken, Brian Chen, Adam Pearce, Craig Citro, Emmanuel Ameisen, Hoagy~Cunningham Andy~Jones, Nicholas~L Turner, Callum McDougall, Monte MacDiarmid, Alex Tamkin, Esin Durmus, Tristan Hume, Francesco Mosconi, C.~Daniel Freeman, Theodore~R. Sumers, Edward Rees, Joshua Batson, Adam Jermyn, Shan Carter, Chris Olah, and Tom Henighan.
\newblock {S}caling {M}onosemanticity: {E}xtracting {I}nterpretable {F}eatures from {C}laude 3 {S}onnet --- transformer-circuits.pub.
\newblock \url{https://transformer-circuits.pub/2024/scaling-monosemanticity/}, 2024{\natexlab{b}}.
\newblock [Accessed 11-02-2025].

\bibitem[Tenenbaum et~al.(2011)Tenenbaum, Kemp, Griffiths, and Goodman]{tenenbaum2011grow}
Joshua~B Tenenbaum, Charles Kemp, Thomas~L Griffiths, and Noah~D Goodman.
\newblock How to grow a mind: Statistics, structure, and abstraction.
\newblock \emph{science}, 331\penalty0 (6022):\penalty0 1279--1285, 2011.

\bibitem[Thasarathan et~al.(2025)Thasarathan, Forsyth, Fel, Kowal, and Derpanis]{thasarathan2025universalsparseautoencodersinterpretable}
Harrish Thasarathan, Julian Forsyth, Thomas Fel, Matthew Kowal, and Konstantinos Derpanis.
\newblock Universal sparse autoencoders: Interpretable cross-model concept alignment, 2025.
\newblock URL \url{https://arxiv.org/abs/2502.03714}.

\bibitem[Thorpe(1989)]{thorpe_local_1989}
Simon Thorpe.
\newblock Local vs. {Distributed} {Coding}.
\newblock \emph{Intellectica. Revue de l'Association pour la Recherche Cognitive}, 8\penalty0 (2):\penalty0 3--40, 1989.
\newblock ISSN 0769-4113.
\newblock \doi{10.3406/intel.1989.873}.
\newblock URL \url{https://www.persee.fr/doc/intel_0769-4113_1989_num_8_2_873}.

\bibitem[Tibshirani(1996)]{tibshirani_regression_1996}
Robert Tibshirani.
\newblock Regression {Shrinkage} and {Selection} {Via} the {Lasso}.
\newblock \emph{Journal of the Royal Statistical Society Series B: Statistical Methodology}, 58\penalty0 (1):\penalty0 267--288, jan 1996.
\newblock ISSN 1369-7412, 1467-9868.
\newblock \doi{10.1111/j.2517-6161.1996.tb02080.x}.
\newblock URL \url{https://academic.oup.com/jrsssb/article/58/1/267/7027929}.

\bibitem[Touvron et~al.(2023)Touvron, Lavril, Izacard, Martinet, Lachaux, Lacroix, Rozière, Goyal, Hambro, Azhar, Rodriguez, Joulin, Grave, and Lample]{touvron_llama_2023}
Hugo Touvron, Thibaut Lavril, Gautier Izacard, Xavier Martinet, Marie-Anne Lachaux, Timothée Lacroix, Baptiste Rozière, Naman Goyal, Eric Hambro, Faisal Azhar, Aurelien Rodriguez, Armand Joulin, Edouard Grave, and Guillaume Lample.
\newblock {LLaMA}: {Open} and {Efficient} {Foundation} {Language} {Models}, feb 2023.
\newblock URL \url{http://arxiv.org/abs/2302.13971}.
\newblock arXiv:2302.13971 [cs].

\bibitem[Treisman(1996)]{treisman1996binding}
Anne Treisman.
\newblock The binding problem.
\newblock \emph{Current opinion in neurobiology}, 6\penalty0 (2):\penalty0 171--178, 1996.

\bibitem[Tsaig and Donoho(2006)]{tsaig2006extensions}
Yaakov Tsaig and David~L Donoho.
\newblock Extensions of compressed sensing.
\newblock \emph{Signal processing}, 86\penalty0 (3):\penalty0 549--571, 2006.

\bibitem[Vafaii et~al.(2024)Vafaii, Galor, and Yates]{vafaii_poisson_2024}
Hadi Vafaii, Dekel Galor, and Jacob~L. Yates.
\newblock Poisson {Variational} {Autoencoder}, may 2024.
\newblock URL \url{http://arxiv.org/abs/2405.14473}.
\newblock arXiv:2405.14473 [cs, q-bio].

\bibitem[Van~Gelder(1990{\natexlab{a}})]{van_gelder_compositionality_1990}
Tim Van~Gelder.
\newblock Compositionality: {A} connectionist variation on a classical theme.
\newblock \emph{Cognitive Science}, 14\penalty0 (3):\penalty0 355--384, 1990{\natexlab{a}}.
\newblock URL \url{https://www.sciencedirect.com/science/article/pii/036402139090017Q}.
\newblock Publisher: Elsevier.

\bibitem[Van~Gelder(1990{\natexlab{b}})]{van_gelder_what_1990}
Tim Van~Gelder.
\newblock What is the ‘{D}’in ‘{PDP}’? {An} overview of the concept of distribution.
\newblock \emph{Philosophy and Connectionist Theory}, pages 33--59, 1990{\natexlab{b}}.

\bibitem[von Helmholtz(1878)]{helmholtz1878}
Herman von Helmholtz.
\newblock \emph{The facts of perception}.
\newblock The Selected Writings of Hermann von Helmholtz, R. Karl, Ed. Middletown, CT, USA: Wesleyan Univ. Press, 1971., 1878.

\bibitem[Vulić et~al.(2017)Vulić, Gerz, Kiela, Hill, and Korhonen]{vulic_hyperlex_2017}
Ivan Vulić, Daniela Gerz, Douwe Kiela, Felix Hill, and Anna Korhonen.
\newblock {HyperLex}: {A} {Large}-{Scale} {Evaluation} of {Graded} {Lexical} {Entailment}.
\newblock \emph{Computational Linguistics}, 43\penalty0 (4):\penalty0 781--835, dec 2017.
\newblock ISSN 0891-2017.
\newblock \doi{10.1162/COLI_a_00301}.
\newblock URL \url{https://doi.org/10.1162/COLI_a_00301}.

\bibitem[Wittgenstein(1953)]{wittgenstein_philosophische_1953}
Ludwig Wittgenstein.
\newblock Philosophische {Untersuchungen}, hg. {GEM} {Anscombe} und {R}. {Rhees}, 1953.

\bibitem[Wright and Sharkey(2024)]{wright2024addressing}
Benjamin Wright and Lee Sharkey.
\newblock Addressing feature suppression in saes.
\newblock In \emph{AI Alignment Forum}, page~16, 2024.

\bibitem[Wu et~al.(2024)Wu, Gao, Dupré~la Tour, and Tillman]{wu2024openai}
Jeffrey Wu, Leo Gao, Tom Dupré~la Tour, and Henk Tillman.
\newblock \emph{Extracting {Concepts} from {GPT}-4}, 2024.
\newblock URL \url{https://openai.com/index/extracting-concepts-from-gpt-4/}.

\bibitem[Xu et~al.(2024)Xu, Yao, Lachapelle, Taslakian, von Kügelgen, Locatello, and Magliacane]{xu_sparsity_2024}
Danru Xu, Dingling Yao, Sébastien Lachapelle, Perouz Taslakian, Julius von Kügelgen, Francesco Locatello, and Sara Magliacane.
\newblock A {Sparsity} {Principle} for {Partially} {Observable} {Causal} {Representation} {Learning}, mar 2024.
\newblock URL \url{http://arxiv.org/abs/2403.08335}.
\newblock arXiv:2403.08335 [cs, stat].

\bibitem[Yun et~al.(2021)Yun, Chen, Olshausen, and LeCun]{yun_transformer_2021}
Zeyu Yun, Yubei Chen, Bruno~A. Olshausen, and Yann LeCun.
\newblock Transformer visualization via dictionary learning: contextualized embedding as a linear superposition of transformer factors, 2021.
\newblock URL \url{http://arxiv.org/abs/2103.15949}.
\newblock arXiv:2103.15949 [cs].

\bibitem[Zhang et~al.(2019)Zhang, Chen, Cheung, and Olshausen]{zhang_word_2019}
Juexiao Zhang, Yubei Chen, Brian Cheung, and Bruno~A. Olshausen.
\newblock Word {Embedding} {Visualization} {Via} {Dictionary} {Learning}, oct 2019.
\newblock URL \url{http://arxiv.org/abs/1910.03833}.
\newblock arXiv:1910.03833 [cs].

\bibitem[Zheng and Zhang(2023)]{zheng_generalizing_2023}
Yujia Zheng and Kun Zhang.
\newblock Generalizing {Nonlinear} {ICA} {Beyond} {Structural} {Sparsity}, nov 2023.
\newblock URL \url{http://arxiv.org/abs/2311.00866}.
\newblock arXiv:2311.00866 [cs, eess, stat].

\bibitem[Zimmermann et~al.(2024)Zimmermann, Klindt, and Brendel]{zimmermann_measuring_2024}
Roland Zimmermann, David Klindt, and Wieland Brendel.
\newblock Measuring {Mechanistic} {Interpretability} at {Scale} {Without} {Humans}.
\newblock \emph{Advances in Neural Information Processing Systems}, 2024.
\newblock URL \url{https://openreview.net/forum?id=M8yBcvRvwn}.

\bibitem[Zimmermann et~al.(2021)Zimmermann, Sharma, Schneider, Bethge, and Brendel]{zimmermann_contrastive_2021}
Roland~S. Zimmermann, Yash Sharma, Steffen Schneider, Matthias Bethge, and Wieland Brendel.
\newblock Contrastive learning inverts the data generating process.
\newblock In \emph{International {Conference} on {Machine} {Learning}}, pages 12979--12990. PMLR, feb 2021.
\newblock URL \url{https://proceedings.mlr.press/v139/zimmermann21a.html}.
\newblock arXiv: 2102.08850.

\end{thebibliography}

\end{document}